\pdfoutput=1

\documentclass[11pt]{article}

\usepackage[final]{acl}

\usepackage{times}
\usepackage{latexsym}
\usepackage{enumitem}
\usepackage{subcaption}

\usepackage[T1]{fontenc}

\usepackage[utf8]{inputenc}
\usepackage{booktabs}
\usepackage{microtype}
\usepackage[most]{tcolorbox}
\usepackage{inconsolata}

\usepackage{graphicx}
\usepackage{xspace}
\usepackage{amsmath}

\newcommand{\model}{HiErarchical, length-RObust machine-influenced text detector}
\newcommand{\modelab}{{\fontfamily{cmss}\fontseries{m}\fontshape{n}\selectfont HERO}}

\makeatletter
\DeclareRobustCommand\onedot{\futurelet\@let@token\@onedot}
\def\@onedot{\ifx\@let@token.\else.\null\fi\xspace}
\def\eg{\emph{e.g}\onedot} \def\Eg{\emph{E.g}\onedot}
\def\ie{\emph{i.e}\onedot}
\def\Ie{\emph{I.e}\onedot}

\makeatother

\title{Real, Fake, or Manipulated? Detecting Machine-Influenced Text}

\author{
 \textbf{Yitong Wang*\textsuperscript{1}} \hspace{4mm}
 \textbf{Zhongping Zhang*\textsuperscript{1}} \hspace{4mm}
 \textbf{Margherita Piana\textsuperscript{1}} \hspace{4mm}
 \textbf{Zheng Zhou\textsuperscript{2}} 
 \\
 \textbf{Peter Gerstoft\textsuperscript{2}} \hspace{4mm}
 \textbf{Bryan A. Plummer\textsuperscript{1}}
\\
 \textsuperscript{1}Boston University,
 \textsuperscript{2}University of California, San Diego
 \\
\small{
    \textbf{Correspondence:} \href{mailto:bplum@bu.edu}{bplum@bu.edu}
  }
}

\newcommand\blfootnote[1]{%
  \begingroup
  \renewcommand\thefootnote{}%
  \footnote{#1}%
  \addtocounter{footnote}{-1}%
  \endgroup
}

\begin{document}
\maketitle

\blfootnote{\hspace{-1.5mm}*Denotes equal contribution}
\begin{abstract}
Large Language Model (LLMs) can be used to write or modify documents, presenting a challenge for understanding the intent behind their use.  For example, benign uses may involve using LLM on a human-written document to improve its grammar or to translate it into another language.  However, a document entirely produced by a LLM may be more likely to be used to spread misinformation than simple translation (\eg, from use by malicious actors or simply by hallucinating).  Prior works in Machine Generated Text (MGT) detection mostly focus on simply identifying whether a document was human or machine written, ignoring these fine-grained uses. In this paper, we introduce a \model{} (\modelab{}), which learns to separate text samples of varying lengths from four primary types: human-written, machine-generated, machine-polished, and machine-translated.  \modelab{} accomplishes this by combining predictions from length-specialist models that have been trained with Subcategory Guidance.  Specifically, for categories that are easily confused (\eg, different source languages), our Subcategory Guidance module encourages separation of the fine-grained categories, boosting performance.  Extensive experiments across five LLMs and six domains demonstrate the benefits of our \modelab{}, outperforming the state-of-the-art by 2.5-3 mAP on average\footnote{Code: \url{https://github.com/ellywang66/HERO}}.
\end{abstract}

\section{Introduction}
\label{sec:intro}

\begin{figure}[t]
  \includegraphics[width=\columnwidth]{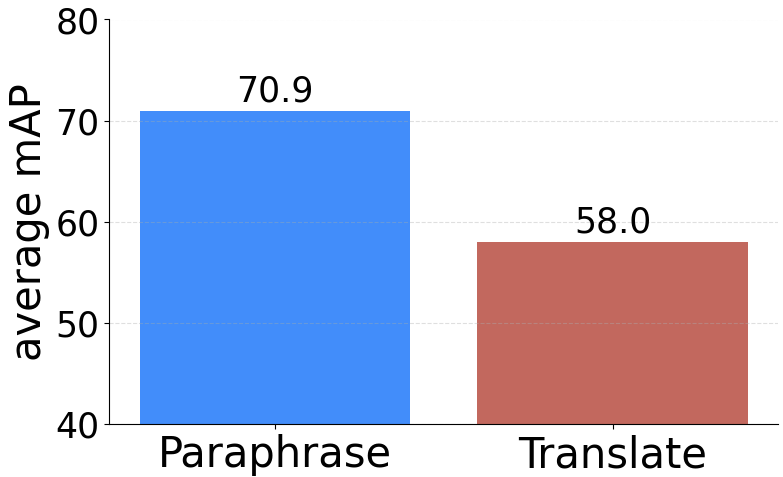}
  \vspace{-8mm}
  \caption{Paraphrasing/polishing human-written text or translating it into another language are often benign applications of an LLM that users of MGT detectors, like moderators, may wish to ignore.  However, off-the-shelf methods (\eg,~\citep{hans2024spotting}) often identify this type of data as machine-generated. In this paper, we increase the practical use of MGT detectors by separating text into fine-grained production categories, providing insight into content intent.}
  \label{fig:random_vs_detector}
\end{figure}

Fine-grained Machine Generated Text (FG-MGT) detection aims to predict if a document was human-written, machine-generated, or some combination thereof. 
 Prior work has primarily focused on separating paraphrased or machine-polished text from human and/or completely machine-generated text~\citep{krishna2024paraphrasing, li2024spotting,abassy2024llm}, as these are often benign uses of a language model.  In contrast, machine generated text may hallucinate~\citep{cao2022hallucinated,parikh2020totto,zhou2021detecting,maynez2020faithfulness,shuster2021retrieval,gou2023critic,meng2022locating} and is more likely to contain misinformation~\citep{lin2022truthfulqa, zellers2019defending}, making them less trustworthy.  However, prior work ignores other benign uses of LLMs, like machine translation, which may also be flagged as machine-generated by traditional MGT detectors (see Fig.~\ref{fig:random_vs_detector}).

\begin{figure}[t]
  \includegraphics[width=\columnwidth]{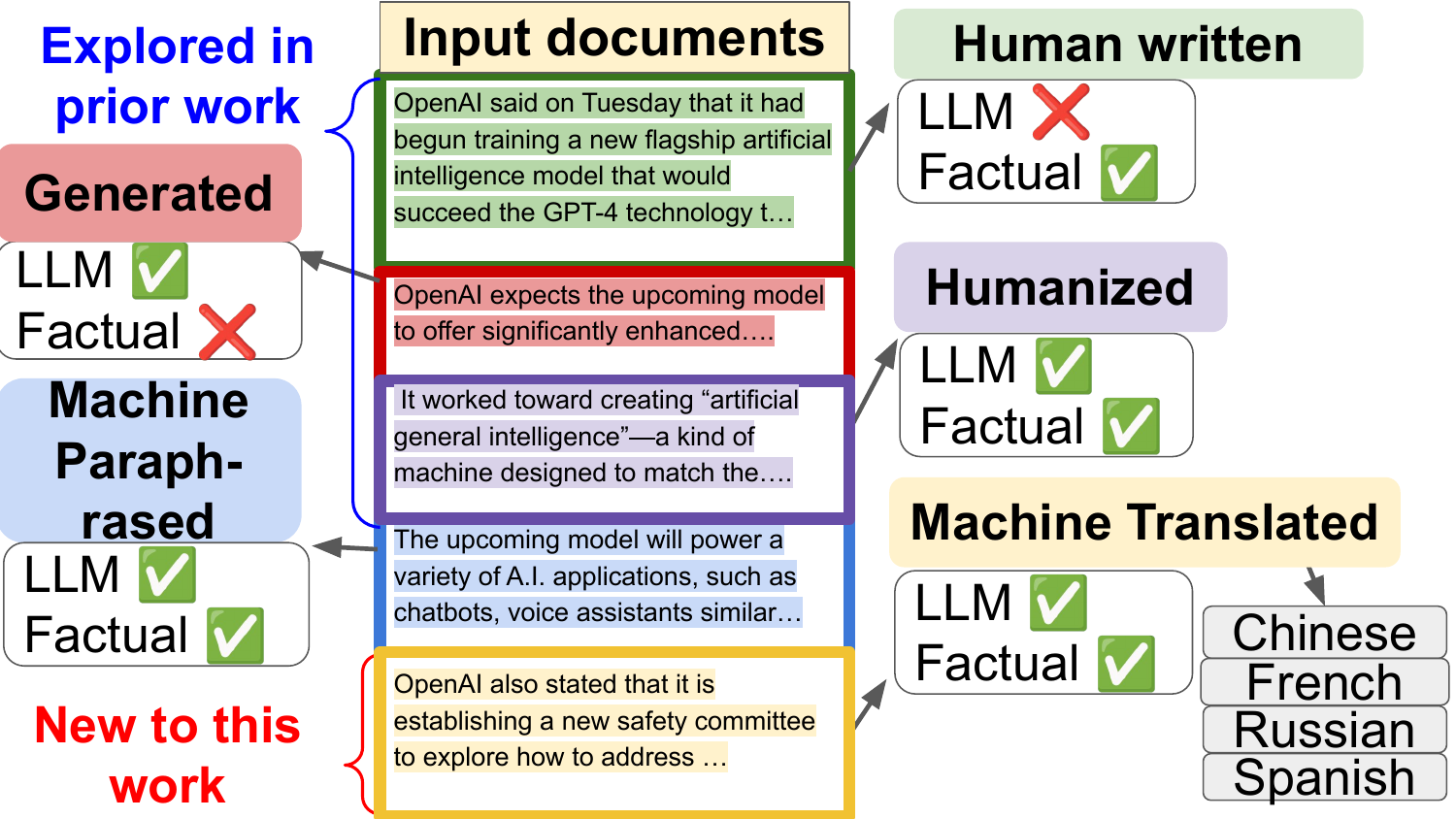}
  \vspace{-6mm}
  \caption{\textbf{Illustration of Fine-grained Machine Generated Text Detection (FG-MGT).} The goal of FG-MGT is to identify different types of generated text to provide some insight into the potential intent. In this paper, we extend the study of \citet{abassy2024llm} to include machine-translated text.}
  \label{fig:task}
\end{figure}

To address this issue, in this paper we introduce a \model{} (\modelab{}), which provides fine-grained labels to better understand document authorship.  Specifically, as shown in Fig.~\ref{fig:task}, we expand the set of possible authorship categories to not only include machine translated text, but also the source language from which it is translated from.  As we will show, identifying the source language both provides more detailed authorship information and also improves the ability of \modelab{} to identify translated text as a whole.   However, separating similar categories of machine-influenced (\ie, translated or polished) text is challenging.  For example, paraphrasing and translation both originate from a human-written article, and a sophisticated actor may also try to make their generated text more human-like in an effort to make it harder to detect.  This is further exacerbated during inference when documents are from different domains or language models than those seen during training. 

A straightforward approach to solve our FG-MGT problem would be to use a coarse-to-fine approach (\eg,~\citep{xu2023dynamic,yuan2023small,amit2004coarse}), where we train a model to predict the general categories, and then refine them using specialized models.   However, this approach has two drawbacks.  First, it can increase inference time as both coarse and fine models must be used for each input document.  Second, it introduces a tradeoff between coarse and fine model predictions that may be challenging to define for strong distribution shifts at test time (\eg, documents from out-of-domain language models).  Thus, as we will show, this type of naive adaptation results in poor performance in practice.  Instead, we introduce Subcategory Guidance modules, where we compute a separate loss function on subsampled logits to distinguish between closely related fine-grained categories. Unlike traditional coarse-to-fine methods, this does not add any additional computational requirements at test time, enabling it to scale to large numbers of categories.

\begin{figure}[t]
\centering
  \includegraphics[width=\linewidth]{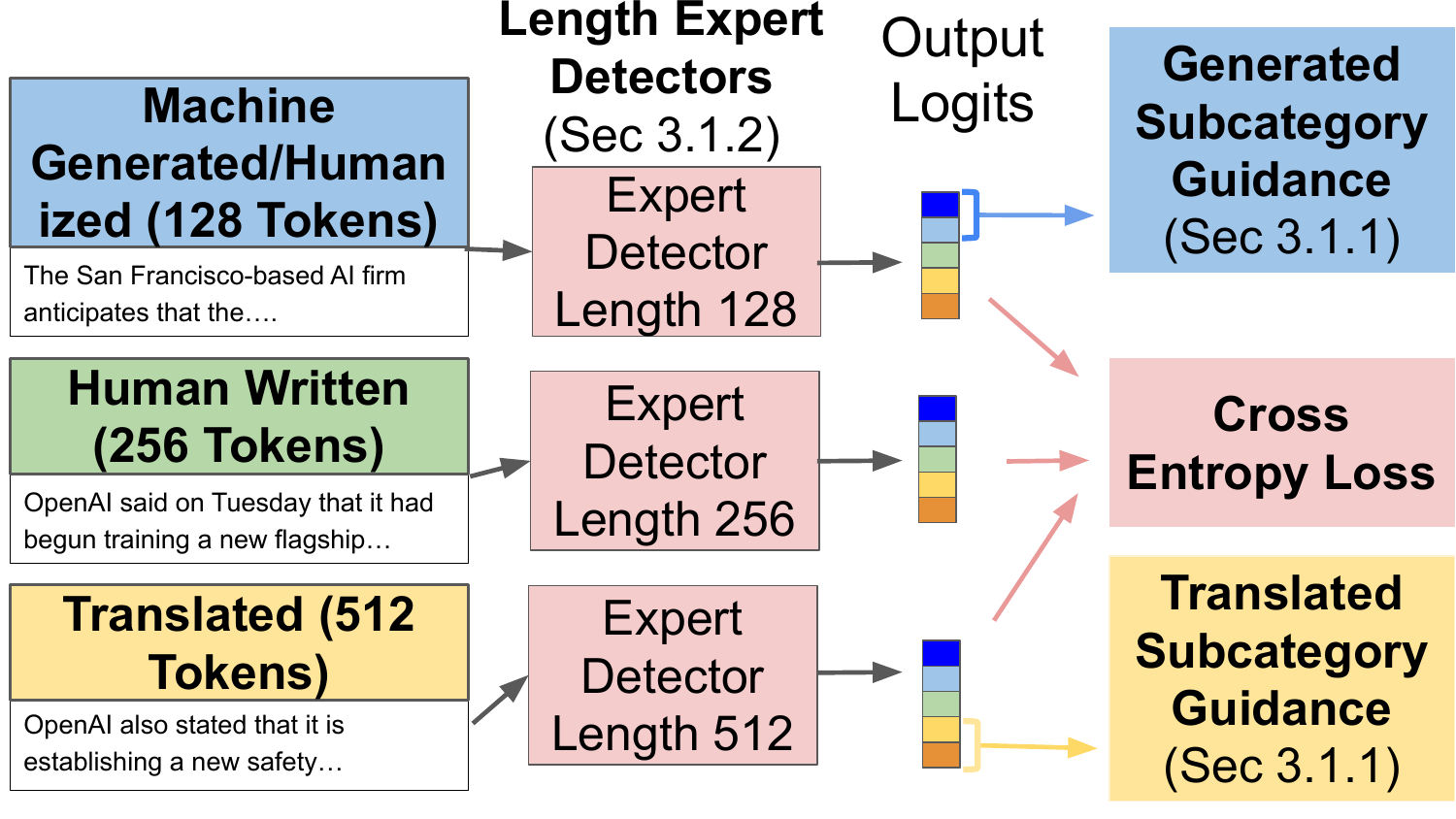}
  \vspace{-8mm}
  \caption{\textbf{Our \modelab{} framework}. Each input is processed by a specialized expert detector based on its token length. In addition to the standard cross-entropy loss, we introduce generated subcategory guidance to machine-generated and machine-humanized text, while translated subcategory guidance is used for translated text. See Sec.~\ref{sec:method} for discussion. }
  \label{fig:model_architecture}
\end{figure}

Another challenge faced in FG-MGT is the variability of input text lengths, where smaller documents prove more challenging to detect.  While this challenge is shared with the traditional MGT task~\citep{hans2024spotting,mitchell2023detectgpt, verma2024ghostbuster, guo2023close, zhang2024MGTL, gehrmann2019gltr, su2023detectllm, tian2023gptzero}, the introduction of fine-grained categories amplifies the issue in our setting.  Inspired by work in bias mitigation~\citep{Wang_2020_CVPR}, we train a set of expert classifiers, each specialized towards a specific text length. Following prior work~\citep{Wang_2020_CVPR}, we use all classifiers at test time regardless of input document length. See Fig.~\ref{fig:model_architecture} for an illustration of our approach.

Our contributions are summarized as follows:
\begin{itemize}[nosep,leftmargin=*]
\item We introduce \modelab{}, a robust FG-MGT detector that combines categories into a hierarchy to focus the model's ability to discriminate between fine-grained categories, which outperforms the state-of-the-art by 2.5-3 mAP on average.
\item We show Subcategory Guidance modules provide an effective approach for separating similar categories without incurring test-time resource costs suffered by related work. 
\item We conduct an in-depth analysis on FG-MGT using \modelab{} to identify potential manipulation and misinformation in text content to ensure the safe deployment of LLMs.

\end{itemize}

\section{Related Work}
\label{sec:related}

Most prior work in detecting Machine Generated Text (MGT) treats this task as a binary classification problem~\citep{bhattacharjee2023conda, solaiman2019release, guo2023close, tian2024multiscale, mitchell2023detectgpt, hans2024spotting, zhang2024MGTL, hu2023radar, kuznetsov-etal-2024-robust}, \ie, detecting whether the input text is human-written or machine-generated. 
These include Metric-based methods~\citep{mitchell2023detectgpt, su2023detectllm, bao2024fast, hans2024spotting,mirallesgonzález2025tokenscreatedequalperplexity}, which extract distinguishable features from the text using the target language models. \Eg, \citet{solaiman2019release} apply log probability, \citet{gehrmann2019gltr} use the absolute rank of each token, and ~\citet{verma2024ghostbuster} searches over a language model's feature space. Many of these methods (\eg,~\citep{mitchell2023detectgpt, su2023detectllm, bao2024fast}, rely on an observation that small changes to generated text typically lower its log probability under the language model, a pattern not seen in human-written text.  Thus, these methods inject perturbations to the input text.  However, these models are only defined for the binary classification, and it is unclear if they can be extended to our setting as we need to separate many types of machine influenced text.

Some recent studies have recognized the importance of detecting other categories of MGT~\citep{krishna2024paraphrasing, li2024spotting, nguyen2021machine}, including machine paraphrased and translated text. For example, \citet{krishna2024paraphrasing} enhanced machine paraphrased text detection using retrieval methods, and~\citet{li2024spotting} identified paraphrased sentences through article context. \citet{nguyen2021machine} applied round-trip translation to detect Google-translated text. \citet{macko-etal-2023-multitude,10888686} explored detecting generated text in non-English languages, but not machine-translated text. \citet{abassy2024llm} explored a fine-grained reason task similar to ours, but did not consider the effect of machine-translated text.  The high similarity between the sub-categories can also reduce the generalization of such an approach to detect other types of manipulations. 
\section{Expanding Fine-grained Machine Generated Text Detection}
\label{sec:method}

Given an article \( x_i \in \mathcal{X} \), fine-grained machine-generated text (FG-MGT) detection aims to separate samples into a set of categories $y_i \in \{0, 1, \dots, K\} $ where \( y_i=0\) corresponds to human-written text, and \( y_i = k \) where \(k \in \{1, \dots, K\} \) corresponds to one of \( K \) distinct categories of machine-influenced text.  Prior work on FG-MGT explored up to four categories: \textbf{human} written, machine \textbf{generated}, \textbf{humanized} machine generated, and \textbf{paraphrased}/polished human written text~\citep{krishna2024paraphrasing, li2024spotting,abassy2024llm}.  However, this ignores \textbf{translated} text, another form of machine-influenced generation with often benign use, but, as shown in Fig.~\ref{fig:random_vs_detector}, may be detected as LLM-generated.  Thus, to provide additional insight for users of FG-MGT models, we add a new category based on the source language a document was translated from.  However, as we will show, we find that separating these types of similar generation types is challenging, especially on out-of-domain generators used at test time.

To address our FG-MGT task, we introduce \model{} (\modelab{}), which makes two improvements to FG-MGT detectors.  First, Sec.~\ref{sec:guided_class} describes our Subcategory Guidance modules, which help construct a feature representation that can more easily separate similar categories.  Second, Sec.~\ref{sec:lengths} discusses our length-expert approach to improving support for varying document lengths.  Sec.~\ref{sec:data} discusses our data generation process that we use to train and evaluate our FG-MGT detectors.

\subsection{Our \modelab{} Approach}

As discussed earlier, our objective is to create a FG-MGT model that can identify if a document is machine-generated and the specific type of machine influence. While our approach is designed to generalize across a wide range of authorship types and languages, in this paper we focus on predicting likelihoods over eight categories for English articles: human written, machine generated, paraphrased, humanized, translated (Chinese), translated (Russian), translated (Spanish), and translated (French) as defined at the beginning of Sec.~\ref{sec:method}. Our HERO model begins by taking our input document $x$ passes it through a shared feature encoder $g$.  To learn to identify our categories above, we use cross entropy $\mathcal{L}_{CE}$, whose classifier uses the input from $g(x)$ and estimates the likelihood that sample $x$ was produced by one of the FG-MGT categories.

A simple approach would be to simply change an MGT detector (\eg,~\citep{hans2024spotting,mitchell2023detectgpt, verma2024ghostbuster, guo2023close, zhang2024MGTL, gehrmann2019gltr, su2023detectllm, tian2023gptzero}) to produce a multi-class outputs.  However, we found these models struggle to distinguish between similar generation types, especially  when evaluated on out-of-distribution language models.  We address this issue with a Subcategory Guidance module in the next section.

\subsubsection{Fine-grained Text Classification via Subcategory Guidance}
\label{sec:guided_class}

One common strategy for discriminating between fine-grained categories is to build a coarse-to-fine hierarchy~\citep{xu2023dynamic,yuan2023small,amit2004coarse}, where categories become more similar as you traverse down the hierarchy.  However, these methods are often deployed within a single domain, \ie, the distribution of the data see during training is similar to that seen at test time.  This is due, in part, to the fact that these methods require careful tuning to balance the predictions of the hierarchy of classifiers being deployed. \Ie, they require careful calibration between the coarse and fine-grained classifiers to boost performance.  In FG-MGT, this would put a significant limitation on our detectors, as it would effectively mean that we can only deploy them on seen text domains and for language models used during training.

Instead, we introduce a Subcategory Guidance module to help direct feature learning during training, which is discarded at test time. We group together semantically similar categories that specialize in separating samples in each group.  Specifically, we create one module for each of the four translated categories as well as for machine-generated and humanized text. Although the machine generated and humanized text are both entirely generated, the fact that a user decided to query a language model to make the text appear more human suggests they might be trying to obfuscate a detector, providing some potential intent information.  Similarly, knowing the language a document was translated from can provide clues as to where a document first appeared.  Our Subcategory Guidance models aim to help our detector better discriminate between these categories.

Unlike the coarse-to-fine methods discussed earlier, these modules are discarded at test time.  Thus, they do not affect computational resources at test time or require complicated calibration procedures that do not generalize well to out-of-domain samples.  Instead, they boost performance by guiding the formation of the shared feature space produced by the shared encoder $g$ during training.  Each Subcategory Guidance module takes as input samples that stem only from the categories of their type.  For example, the Translated Subcategory Guidance only takes features from documents from the four translated categories as input.  Then it uses cross entropy to separate documents into their fine-grained categories.  In effect, this simply amounts to computing a loss over a subset of predictions, making it easy to implement and deploy.

Our final objective consists of a tradeoff function balancing the task loss with our Subcategory Modules, which we define as $\mathcal{L}_{\text{GH}}$ and $\mathcal{L}_{\text{Trans}}$ for the generated/humanized and translated categories, respectively.  Formally, our total loss is:
\begin{equation}  
\mathcal{L}_{\text{Total}} = \mathcal{L}_{\text{CE}} + \lambda(\mathcal{L}_{\text{GH}} + \mathcal{L}_{\text{Trans}}),
\label{eq:guided}
\end{equation}
where $\lambda$ is a tunable hyper-parameter.

\subsubsection{Improving Support to Varying Document Lengths}
\label{sec:lengths}

Prior work has shown that short documents, which inherently have little information about authorship, are challenging to identify as machine generated~\citep{zhang2024MGTL}. \citet{solaiman2019release} found they could improve a detector's robustness to varying document lengths by randomly cropping articles during training.  However, a detector for short length article has to naturally be more sensitive to distribution changes given the limited information than it does for a longer article.  Training a single model to adjust for both the sensitivity as well as make fine-grained distinctions is challenging.  Instead, we leverage a set of experts, each of which specializes in documents up to a set length.

Formally, given an input text \( \mathbf{x} \), we train a set of \( M \) expert classifiers \( \{ f_1, \dots, f_M \} \), each trained with a specific maximum token length and associated parameters \( \mathbf{W}_m \). Each expert is trained using Subcategory Guidance from Sec.~\ref{sec:guided_class}.  However, empirically we find that including some information from documents of lengths other than the ones targeted by an expert can help improve performance (\eg, seeing some 256 token length documents can boost performance for a 512-length expert).  Thus, we used length cropping, where with $p_{\text{crop}}$, documents of other lengths are included during training to improve the model's robustness.  

Given a document at test time we can simply use the expert of the closest length.  If a document is between experts, we use the larger one.  However, some prior work in bias mitigation has shown that averaging experts even over settings they do not specialize in can boost performance~\cite{Wang_2020_CVPR}.  In effect, when compute is available, these experts can form a type of ensemble.  Thus, in our experiments we evaluate these experts as an ensemble in addition to using them individually.

\subsection{Data Preparation: Article Generation}
\label{sec:data}

We generate articles for a range of domains (Sec.~\ref{sec:source_data}) and language models (Sec.~\ref{sec:generation}) to ensure FG-MGT methods generalize across many settings, which we discuss in more detail below.  

\subsubsection{Source Datasets}
\label{sec:source_data}

\noindent \textbf{GoodNews}~\citep{biten2019good}  
provides URLs of New York Times articles from 2010 to 2018. After filtering out broken links and non-English articles, we randomly selected 8K/2K/2K articles for train/test/validation splits.
\smallskip

\noindent \textbf{VisualNews}~\citep{liu2020visualnews} has articles from four media sources: \emph{Guardian}, \emph{BBC}, \emph{USA Today}, and \emph{Washington Post}. We randomly selected 2K articles for evaluation.
\smallskip

\noindent \textbf{Student essays (Essay), creative writing (WP), and news articles (Reuters)}~\citep{verma2024ghostbuster} represent three diverse domains with 1K articles from each dataset used for evaluation.
\smallskip

\noindent \textbf{WikiText}~\citep{merity2016pointer} contains 60 test articles collected from Wikipedia.
\smallskip

Each source dataset above provides per-category article counts. \Eg, for GoodNews this results in 2K human written articles * 8 categories = 16K total test samples.  Additionally, only GoodNews was used for training, so results on the remaining datasets provide insight into domain shifts of varying degrees (\eg, VisualNews and Reuters being close domains, whereas the rest are far domains).

\subsubsection{Generation Process}
\label{sec:generation}

To ensure the quality of generated text we keep all prompts for each category consistent throughout the generation process. All other hyperparameters such as temperature for each LLM are also kept the same for consistency. Specifically, the language models we used include Llama-3~\citep{touvron2023llama}, Qwen-1.5~\citep{bai2023qwen}, StableLM-2~\citep{bellagente2024stable}, ChatGLM-3~\citep{du2022glm}, and Qwen-2.5~\citep{yang2024qwen2}. Llama-3 is set as our in-domain generator used for training the detector, and StableLM-2, ChatGLM-3, Qwen-1.5, and Qwen-2.5 are out-of-domain generators to evaluate the model's generalization ability. When generating articles, we used a temperature for Llama-3 of 0.6 and for StableLM-2 we used 0.7. For Qwen-1.5, Qwen-2.5, and ChatGLM-3 we use default temperatures for generating responses.

To prevent the model from leaking information about the article's category (\eg, Llama-3 often responds with "Here is the polished version:"), we use the text starting from the second sentence as input to the detector.  
Below we further discuss category-specific generation processes.
\smallskip

\noindent\textbf{Machine-generated} articles were created by giving the LLM the title with the prompt:  ``Write an article on the following title, ensuring that the article consists of approximately $z$ sentences," where $z$ represents the number of sentences in the original article. This ensures that articles of different categories are of similar length, preventing the detector from using length as a classification feature.
\smallskip

\noindent\textbf{Machine-paraphrased} articles were generated by giving the LLM the entire human-written article as input with the prompt "Paraphrase the following article: $x$."  We provide a study the effect of replacing only parts of the articles in App.~\ref{sec:appendix_paraphrase}.
\smallskip

\noindent\textbf{Machine-translated} articles were produced using the same process as for paraphrasing, only replacing the word "paraphrase" with "translate" in the prompt. Translated articles were drawn from the following languages: Chinese, Spanish, Russian, and French (additional discussion in App.~\ref{sec:appendix_round-trip_translation}).
\smallskip

\noindent\textbf{Machine-humanized} articles were created by giving the a machine-generated article as input to the LLM with the prompt: “Rewrite this text to make it sound more natural and human-written.” We provide a specific example in App.~\ref{sec:appendix_humanized}.

\begin{table*}[h!]
\centering
\setlength{\tabcolsep}{2.pt}
\begin{tabular}{l cccccccc}
\toprule
& \multicolumn{1}{c}{In-domain LLMs} & \multicolumn{4}{c}{Out-of-domain LLMs} & \\
\cmidrule(r){2-2} \cmidrule(r){3-7}
\textbf{Model} & Llama3 & Qwen1.5 & Qwen2.5 & ChatGLM3 & StableLM2 & \textbf{avg} &  \textbf{PD} \\
Scale & -8B & -7B & -12B & -6B & -7B & \textbf{mAP} & \textbf{5\%FPR}\\
\midrule
\multicolumn{8}{c}{\textbf{In-domain mAP on GoodNews~\citep{biten2019good}}} \\
\midrule
OpenAI-D (large) & 94.04 & 41.60 & \textbf{41.90} & 48.23 & 71.21 & 59.39 & 53.31 \\
ChatGPT-D & 80.83 & 38.79 & 40.43 & 40.00 & 62.74 & 52.56 &47.08\\
LLM-DetectAIve & 96.24 & 41.27 & 42.28 & 44.72 & 76.87 & 60.28 & 55.31\\
DistilBERT & 96.89 & 38.99 & 40.91 & 42.21 & 74.59 & 58.72 &53.77\\
\modelab{} (ours) & \textbf{98.33} & \textbf{44.05} & 41.88 & \textbf{50.47} & \textbf{76.93} & \textbf{62.33} &\textbf{56.23}\\
\midrule
\multicolumn{8}{c}{\textbf{Out-of-domain mAP on VisualNews~\citep{liu2020visualnews}}} \\
\midrule
OpenAI-D (large) & 60.67 & 32.62 & \textbf{37.99} & 38.51 & 52.53 & 44.46 &36.64 \\
ChatGPT-D & 47.19 & 27.39 & 31.02 & 33.82 & 49.93 & 37.87 &31.20\\
LLM-DetectAIve & 62.41 & 32.54 & \textbf{39.41} & 36.34 & \textbf{55.70} & 45.28 & 38.51 \\
DistilBERT & 61.11 & 31.64 & 32.77 & 36.75 & 54.43 & 43.34 &36.49\\
\modelab{} (ours) & \textbf{64.17} & \textbf{38.98} & 37.70 & \textbf{42.17} & 55.48 & \textbf{47.70} &\textbf{39.09} \\
\midrule
\multicolumn{8}{c}{\textbf{Out-of-domain mAP on WikiText~\citep{merity2016pointer}}} \\
\midrule
OpenAI-D (large) & 65.39 & 33.65 & \textbf{36.43} & 36.38 & 52.06 & 44.78 &35.33 \\
ChatGPT-D & 38.78 & 29.16 & 32.76 & 30.78 & 43.08 & 34.91 &24.38\\
LLM-DetectAIve & 65.62 & 29.89 & 28.55 & 27.52 & 45.38 & 39.39 & 29.92\\
DistilBERT & 66.37 & 33.51 & 28.42 & 30.19 & 49.82 & 41.66 &33.12\\
\modelab{} (ours) & \textbf{72.19} & \textbf{37.97} & 31.52 & \textbf{35.45} & \textbf{52.58} & \textbf{45.94} & \textbf{37.00}\\
\midrule
\multicolumn{8}{c}{\textbf{Out-of-domain mAP on WP~\citep{he2023mgtbench}}} \\
\midrule
OpenAI-D (large) & 55.48 & 46.69 & 44.88 & 37.44 & \textbf{55.34} & 47.97 &23.50\\
ChatGPT-D & 40.57 & 37.71 & 42.22 & 34.86 & \textbf{43.69} & 39.81 &17.50\\
LLM-DetectAIve & 65.39 & 48.53 & \textbf{51.35} & 35.35 & \textbf{56.32} & 51.39 & 28.50\\
DistilBERT & 71.65 & 44.61 & 50.88 & 40.51 & 49.74 & \textbf{51.48} &29.25\\
\modelab{} (ours) & \textbf{73.68} & 47.14 & 41.01 & 39.12 & 52.58 & 50.71 &\textbf{42.55}\\
\midrule
\multicolumn{8}{c}{\textbf{Out-of-domain mAP on Reuters~\citep{he2023mgtbench}}} \\
\midrule
OpenAI-D (large) & 74.63 & \textbf{50.72} & \textbf{51.51} & 54.11 & 54.08 & 57.01 &26.75\\
ChatGPT-D & 57.42 & 46.03 & 48.28 & 50.65 & 44.05 & 49.29 &22.00\\
LLM-DetectAIve & \textbf{85.92} & 38.48 & 43.37 & 53.54 & 59.65 & 56.19 &22.75\\
DistilBERT & 81.04 & 48.66 & 42.95 & 42.14 & \textbf{65.93} & 56.14 &31.00\\
\modelab{} (ours) & 84.50 & 48.59 & 41.99 & 50.85 & 58.94 & \textbf{56.98} & \textbf{49.70}\\
\midrule
\multicolumn{8}{c}{\textbf{Out-of-domain mAP on Essay~\citep{he2023mgtbench}}} \\
\midrule
OpenAI-D (large) & 51.29 & 29.09 & 31.74 & \textbf{40.79} & 36.42 & 37.87 &18.25\\
ChatGPT-D & 32.27 & 33.24 & 29.97 & 30.62 & 25.45 & 30.31 &11.75\\
LLM-DetectAIve & 52.56 & 40.22 & \textbf{41.37} & 34.78 & 37.89 & \textbf{41.38} &23.75\\
DistilBERT & 48.98 & \textbf{36.13} & 31.16 & 28.39 & 36.57 & 36.25 &17.25\\
\modelab{} (ours) & \textbf{60.07} & 38.20 & 35.90 & 33.11 & 35.69 & 40.59 &\textbf{33.76}\\
\bottomrule
\end{tabular}
\vspace{-2mm}
\caption{Fine-grained MGT detection results on in-domain GoodNews data and five out-of-domain datasets. \modelab{} outperforms or obtains similar results to prior work in nearly all settings and metrics, providing an overall advantage.}
\label{table:merged_mgt_main}
\end{table*}

\section{Experiments}
\label{sec:experiments}

\noindent\textbf{Implementation Details.}
Our base encoder uses a DistilBERT~\citep{distilbert} backbone. The maximum token length of the input text is set to 512 when training all methods (including our own). The same maximum length is used evaluate the model's performance during testing except where noted.  For training, we used the Adam optimizer with a maximum learning rate of $10^{-5}$.  Following~\citet{zhang2024MGTL,verma2024ghostbuster}, we fine-tuned the model for three epochs to prevent overfitting. Our experiments were conducted on a single GPU (\eg, A40, L40S). For a single dataset (\eg, GoodNews), data preparation takes approximately 60 hours, and training takes around 1 hour. 

\noindent \textbf{Metrics.}  
Our main results use mean Average Precision (mAP) and the probability of detection (PD) at 5\% false positive rate (FPR).  We report mAP per generator, and then rank a detector's overall performance by averaging mAP across both in-domain and out-of-domain LLMs (avg mAP).  We also report F1 score in some ablations.

\subsection{Baselines}
\noindent \textbf{OpenAI-D}~\citep{solaiman2019release} is a detector trained on outputs from GPT-2~\citep{radford2019language} series. OpenAI provides two versions: RoBERTa-base and RoBERTa-large. With fine-tuning and early stopping, OpenAI-D can also be used to detect text generated by other LLMs.
\smallskip

\noindent \textbf{ChatGPT-D}~\citep{guo2023close} is designed to identify text produced by ChatGPT-3.5~\citep{ouyang2022training}. It is trained using the HC3~\citep{guo2023close} dataset, which includes 40,000 questions along with both human-written and ChatGPT-generated answers, before finetuning on our task.
\smallskip

\noindent \textbf{LLM-DetectAIve}~\citep{abassy2024llm} distinguishes between machine-generated, machine-paraphrased, and human-written text by fine-tuning RoBERTa~\citep{liu2019roberta} and DeBERTa~\citep{he2021deberta} models. We apply the DeBERTa backbone of LLM-DetectAIve in our experiments.
\smallskip

\noindent \textbf{DistilBERT}~\citep{distilbert} is a distilled version of BERT~\citep{bert}. Since the model is pre-trained using knowledge distillation, it is smaller and faster at inference time.

\subsection{Results}
\label{sec:experiments_finegrainedMGT}

Tab.~\ref{table:merged_mgt_main} compares the performance of our \modelab{} approach to prior work on in-domain and out-of-domain datasets, respectively.  Note that both tables have results on in-domain and out-of-domain LLMs used for generation. \modelab{} achieves a boost in both mAP and PD 5\%FPT in nearly all settings (\eg, a 2\% boost in avg mAP on GoodNews as seen in Tab.~\ref{table:merged_mgt_main}).  In cases where we underperform prior work on avg mAP, \eg, 1 point worse on WP and Essay in Tab.~\ref{table:merged_mgt_main}, we greatly outperform in PD 5\%FPT (a 12.5\% and 10\% gain, respectively).  Thus, our approach demonstrates significant benefits over the methods from prior work.

\begin{figure*}[t]
    \centering
    \begin{subfigure}[b]{0.49\linewidth}
        \includegraphics[width=\textwidth]{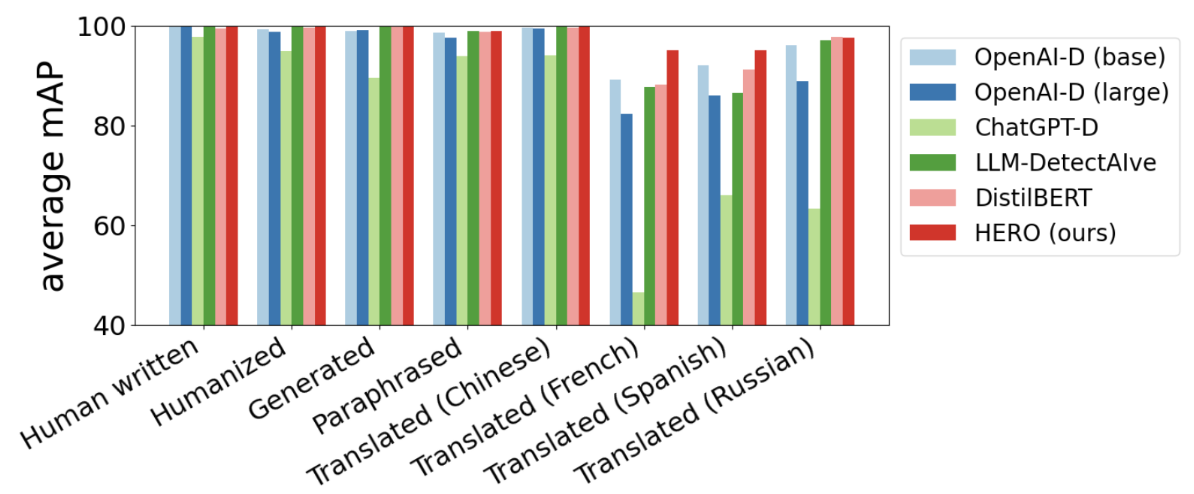}
        \caption{GoodNews In-Domain LLMs}
    \end{subfigure}
    \hfill
    \begin{subfigure}[b]{0.49\linewidth}
        \includegraphics[width=\textwidth]{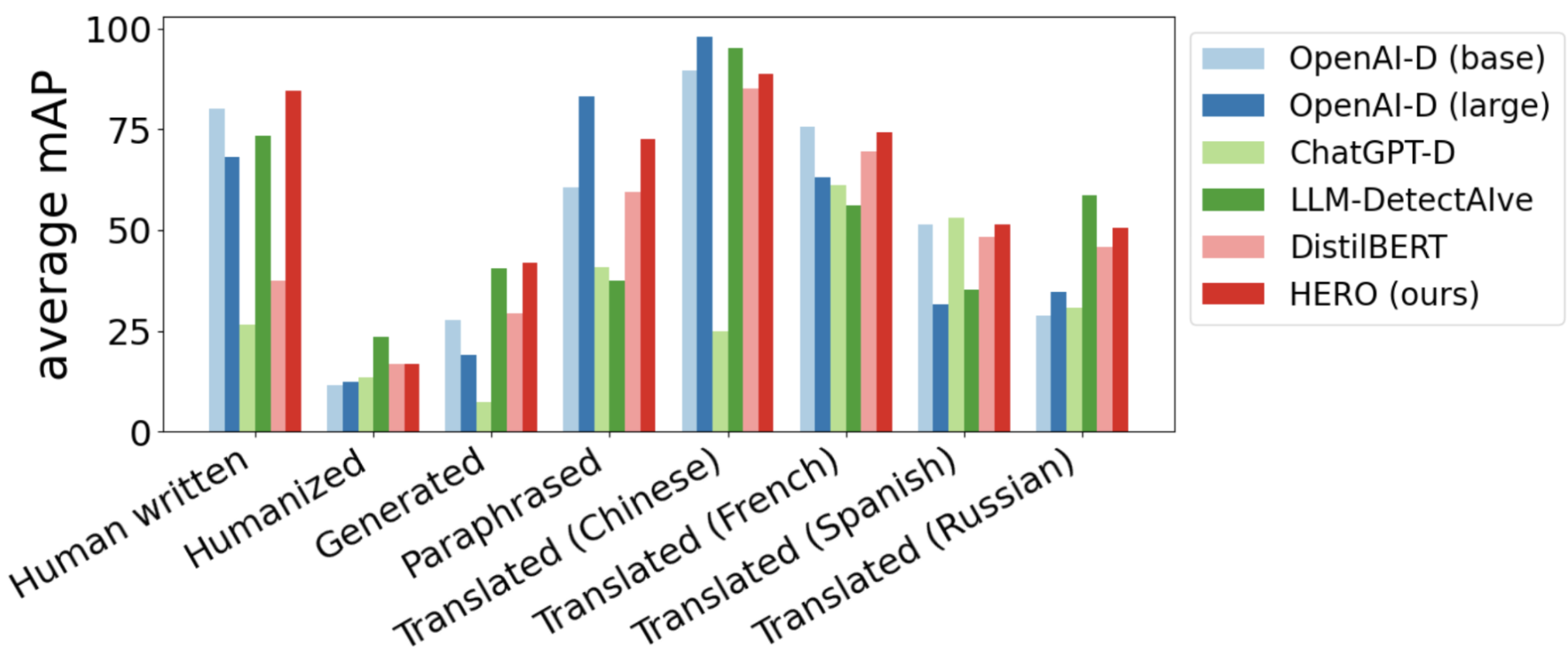}
        \caption{Reuters In-Domain LLMs}
    \end{subfigure}
    
    \begin{subfigure}[b]{0.49\linewidth}
        \includegraphics[width=\textwidth]{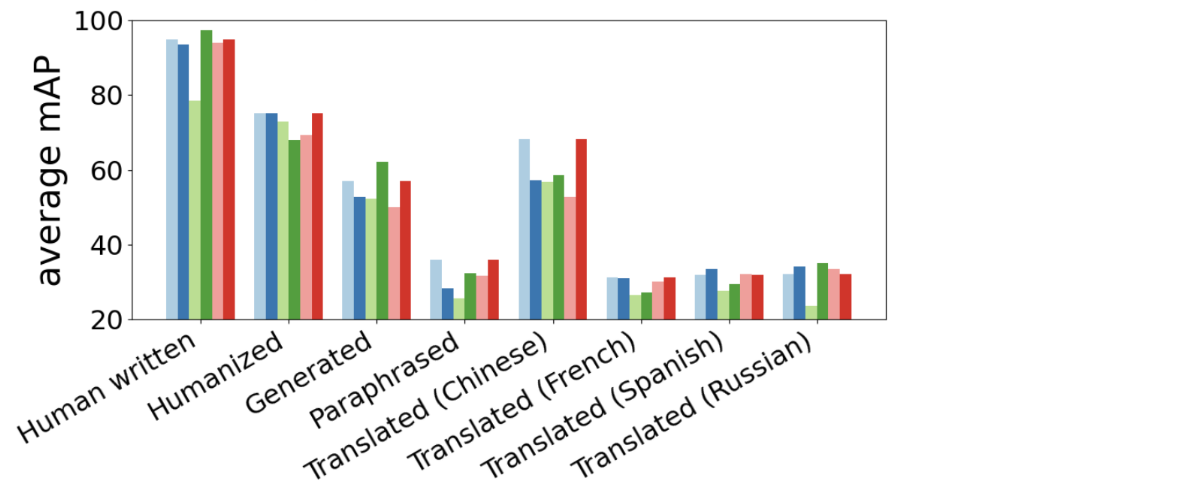}
        \caption{GoodNews Out-of-Domain LLMs}
    \end{subfigure}
    \hfill
    \begin{subfigure}[b]{0.49\linewidth}
        \includegraphics[width=\textwidth]{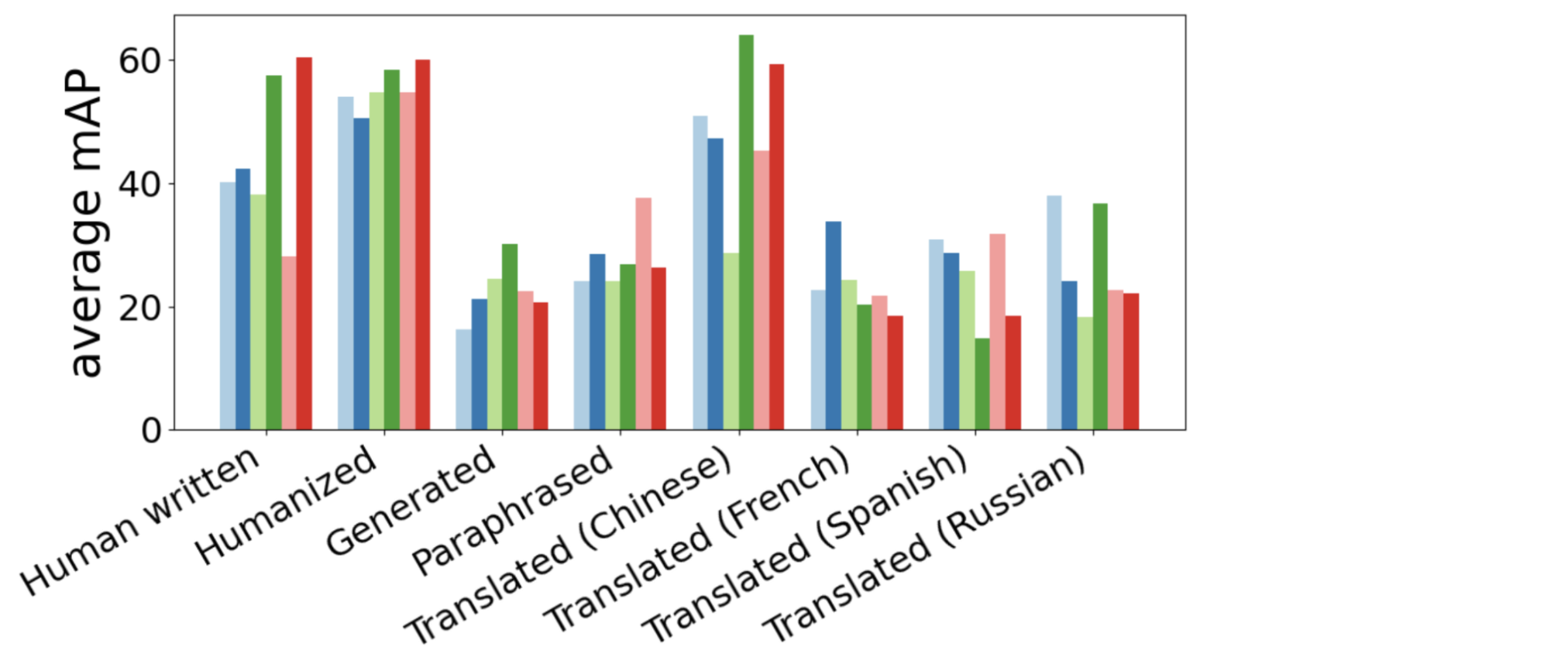}
        \caption{Reuters Out-of-Domain LLMs}
    \end{subfigure}
    
    \caption{Per-class mAP results on the GoodNews~\citep{biten2019good} and Reuters~\citep{verma2024ghostbuster} datasets. 
    Top row: In-domain LLMs. Bottom row: Out-of-domain LLMs. 
    Our method shows more robust performance, especially on human-written and translated categories.}
    \label{fig:per_class}
\end{figure*}

Fig.~\ref{fig:per_class} reports per-class performance on GoodNews and Reuters as representatives of in-domain and out-of-domain data, respectively.  We make two major observations about these results.  First, in-domain data and LLMs gets nearly perfect performance, highlighting the significant role shifts in both has on performance.  For example,  while nearly all methods get perfect performance on humanized data in Fig.~\ref{fig:per_class}(a), when we shift domains (but not LLMs) in Fig.~\ref{fig:per_class}(b) performance drops significantly.  Second, no method gets the best performance consistently across all categories.  For example, while \modelab{} gets best performance identifying out-of-domain human-written articles, LLM-DetectAIive performs best on identifying Russian-sourced translations on Reuters (but performs relatively poorly on paraphrased data).  Thus, our performance improvements come from having more consistent results rather than being strictly better for all categories.

Tab.~\ref{table:token_lengths_analysis} reports performance on various input document lengths using our FG-MGT detectors.  Across all token length settings, performance generally improves with longer token lengths with the best results consistently observed at 500 and 512 tokens. Compared to DistilBERT~\citeyearpar{distilbert}, both the individual length specialist and \modelab{} demonstrate improved performance. The Length Specialist approach shows especially strong performance on short lengths, with the single specialists outperforming the ensemble, validating that such documents require special care.

\begin{table}[t]
\centering
\setlength{\tabcolsep}{1.2pt}
\begin{tabular}{l cccccc}
\toprule
 & Llam & Qwen & StableL & ChatGL & Qwen & \textbf{avg}  \\
 & a3-8B & 1.5-7B & M2-12B & M3-6B & 2.5-7B &  \textbf{mAP}\\
\midrule
\multicolumn{7}{l}{\textbf{(a) DistilBERT~\citeyearpar{distilbert}}} \\
L=32 & 37.97 & 25.38 & 22.75 & 24.10 & 26.42 & 27.33  \\

L=50 & 45.49 & 31.01 & 27.47 & 28.05 & 31.52 & 32.71  \\
L=128  & 58.09 & 37.97 & 33.98 & 33.68 & 39.64 & 40.67 \\
L=256 & 46.41 & 29.46 & 37.60 & 32.01 & 34.19 & 35.93  \\
L=500 & 66.71 & 32.12 & 58.50 & 59.37 & 32.47 & 49.84 \\
L=512 & 61.13 & 31.18 & 54.49 & 35.02 & 32.88 & 42.94 \\
\midrule
\multicolumn{7}{l}{\textbf{(b) \modelab{} (ours) - Single Length Specialist Only}} \\
L=32 & 37.51 & 32.15 & 31.18 & 26.22 & 29.61 & 31.33 \\
L=50  & 44.62 & 36.68 & 34.08 & 30.04 & 33.72 & 35.83 \\
L=128   & 62.23 & 43.94 & 38.53 & 36.81 & 43.79 & 45.06 \\
L=256  & 50.98 & 33.75 & 43.22 & 34.67 & 39.25 & 40.37 \\
L=500 & 63.79 & 37.73 & 56.44 & 40.53 & 38.42 & 47.38 \\
L=512 & 60.87 & 35.36 & 54.51 & 38.00 & 35.50 & 44.85 \\
\midrule
\multicolumn{7}{l}{\textbf{(c) \modelab{} (ours) - All Length Specialists}} \\
L=32 & 36.39 & 31.78 & 30.92 & 25.79 & 29.22 & 30.82 \\
L=50   & 42.65 & 35.75 & 33.66 & 29.04 & 32.93 & 34.81 \\
L=128   & 60.23 & 42.51 & 37.20 & 35.01 & 42.08 & 43.40 \\
L=256 & 51.92 & 32.50 & 39.27 & 34.46 & 38.63 & 39.36 \\
L=500 & 69.91 & 47.39 & 58.98 & 38.47 & 47.07 & 52.36 \\
L=512 & 64.07 & 38.22 & 56.73 & 41.00 & 38.04 & 47.61 \\
\bottomrule
\end{tabular}
\caption{Comparison of mAP scores on VisualNews~\citep{liu2020visualnews} across different input lengths for DistilBERT~\citeyearpar{distilbert} and \modelab{}. \modelab{} consistently outperforms DistilBERT across all lengths and generators. For length-specialist models, we use the expert closest in length, defaulting to the longer one when in between.}
\label{table:token_lengths_analysis}
\end{table}

\begin{table}[t]
\centering
\small
\setlength{\tabcolsep}{1.5pt}
\begin{tabular}{l ccc ccc}
\toprule
& \multicolumn{3}{c}{In-domain LLMs} & \multicolumn{3}{c}{Out-of-domain LLMs} \\
\cmidrule(r){2-4} \cmidrule(r){5-7}
\textbf{Model} & Low & Median & High & Low & Median & High \\
\midrule
\multicolumn{7}{c}{\textbf{In-domain mAP on GoodNews~\citep{biten2019good}}} \\
\midrule
OpenAI-D (base)   & 95.23 & 94.53 & 92.71 & 52.58 & \textbf{27.30} & 29.06 \\
OpenAI-D (large)  & 92.36 & 90.87 & 84.25 & 58.44 & 25.97 & 28.16 \\
ChatGPT-D         & 72.81 & 69.87 & 65.68 & 46.41 & 26.40 & 25.74 \\
LLM-DetectAIve    & 93.45 & 95.03 & 89.59 & \textbf{65.90} & 25.99 & \textbf{31.86} \\
DistilBERT        & 94.85 & 95.49 & 92.46 & 62.97 & 26.55 & 27.38 \\
HERO (ours)       & \textbf{97.41} & \textbf{97.36} & \textbf{95.50} & 64.61 & 26.44 & 27.31 \\
\midrule
\multicolumn{7}{c}{\textbf{Out-of-domain mAP on WikiText~\citep{merity2016pointer}}} \\
\midrule
OpenAI-D (base)   & 67.02 & 69.39 & 64.81 & 31.29 & 29.40 & 32.49 \\
OpenAI-D (large)  & 72.88 & 69.68 & 67.56 & \textbf{42.19} & 30.92 & 33.11 \\
ChatGPT-D         & 53.40 & 44.53 & 40.96 & 28.55 & 29.01 & 27.66 \\
LLM-DetectAIve    & 67.20 & 71.23 & 63.40 & 37.48 & 31.11 & 31.52 \\
DistilBERT        & 70.50 & 65.74 & 57.95 & 34.24 & 29.04 & 32.18 \\
HERO (ours)       & \textbf{73.98} & \textbf{69.19} & \textbf{63.08} & 36.76 & \textbf{32.06} & \textbf{34.13} \\
\bottomrule
\end{tabular}
\caption{Fine-grained MGT detection results by BLEU translation quality. We find that \modelab{} performs especially well across quality groups on out-of-domain data.}
\label{table:translation_quality}
\end{table}

\subsection{\modelab{} Model Analysis}

\begin{table*}[t]
\centering
\small
\setlength{\tabcolsep}{2pt}
\begin{tabular}{l cccccccccc}
\toprule
& \multicolumn{1}{c}{In-domain LLMs} & \multicolumn{4}{c}{Out-of-domain LLMs} &  \\
\cmidrule(r){2-2} \cmidrule(r){3-9}
\textbf{Model} & Llama3 & Qwen1.5 & Qwen2.5 & ChatGLM3 & StableLM2 & \textbf{avg} &  \textbf{PD} &\textbf{F1}\\
Scale & -8B & -7B & -12B & -6B & -7B & \textbf{mAP} & \textbf{5\%FPR} &  \textbf{Score} \\
\midrule
DistilBERT~\citeyearpar{distilbert}& 61.11 & 31.64 & 54.43 & 36.75 & 32.77 & 43.34  &36.49 &32.30 \\ 
\hspace{3mm}+Naive Coarse-to-Fine &42.62 &29.12 &28.36 &26.89 &30.66 &31.53 & 10.67 &22.62 \\
\hspace{3mm}+Subcategory Guidance   & 61.20 & 37.09 & 54.83 & 39.16 & 37.56 & 45.97 &39.09 & 33.87 \\

\hspace{3mm}+Length Cropping~\citeyearpar{solaiman2019release} & 60.69 & 38.24 & 53.84 & 40.06 & 34.20 & 45.40  & 37.66 &31.22 \\
\hspace{3mm}+Length Specialists & 63.21 & 34.44 & 54.05 & 39.72 & \textbf{38.10} & 45.90 & 38.86 &  33.85 \\
\modelab{} (ours)        & \textbf{64.17} & \textbf{38.98} & \textbf{55.48} & \textbf{42.17} & 37.70 & \textbf{47.70} & \textbf{39.09} & \textbf{33.99} &\\ 
\bottomrule
\end{tabular}
\caption{Ablation Study on Visualnews~\citep{liu2020visualnews}. Each component contributes to model performance.  Additionally, our Subcategory Guidance outperforms alternatives like a Naive Coarse-to-Fine approach.}
\label{table:ablations}
\end{table*}

Tab.~\ref{table:ablations} provides an ablation study to show the contribution of each component of \modelab.  We see Subcategory Guidance provides a 2.5 average mAP gain over the baseline DistilBERT~\citep{distilbert}.  We also compare to a naive coarse-to-fine approach that first tries to predict if an input document is human-written, machine-generated, paraphrased, or translated.  If it is machine-generated or translated, we use a separate detector to separate it into the subcategories. Comparing the 2nd and 3rd row of Tab.~\ref{table:ablations}, we see the naive approach underperforms our Subcategory Guidance approach by 14.5 average mAP, highlighting the challenges of generalizing beyond the training domain in our task.  We also show that Length Cropping and our expert models from Sec.~\ref{sec:lengths} both individually boost performance, but when we combine all components we see the best performance.

\begin{table*}[t]
\centering
\setlength{\tabcolsep}{2.5pt}
\begin{tabular}{l ccccccc}
\toprule
\textbf{Model} & Llama3 & Qwen1.5 & StableLM2 & ChatGLM3 & Qwen2.5 & \textbf{avg mAP} & \\
Scale & -8B & -7B & -12B & -6B & -7B &  \\
\midrule
DistilBERT~\citeyearpar{distilbert} &49.58 &45.24 &41.66 &44.30 &45.68 &45.29\\
\modelab{} (Ours)  &\textbf{52.46} &\textbf{47.29} &\textbf{45.59} &\textbf{47.23} &\textbf{48.19} &\textbf{48.15}\\
\bottomrule
\end{tabular}
\caption{Comparison of mAP scores on VisualNews~\citep{liu2020visualnews} for DistilBERT~\citeyearpar{distilbert} and \modelab{} on human-written, machine-generated, machine paraphrased, and machine translated categories.  Identifying source languages can still boost performance even when all translations are treated as a single category.}
\label{table:four_class_analysis}
\end{table*}

Tab.~\ref{table:translation_quality} reports the effect of translation quality binned into low, medium, and high based on BLEU scores. Similar to our per-category results discussed earlier, \modelab{}'s benefits stem from performing better across the varying degrees of translation quality. Notably, our approach performs especially well on out-of-domain data (WikiText results), obtaining the best performance in all but one setting.  We also evaluate how the extent of paraphrasing used affects performance in App.~\ref{sec:appendix_paraphrase}, where \modelab{} typically reports at least a 2 average mAP gain over the baseline DisilBERT model.  These results demonstrate our approach's robustness to a wide range of applications.
\smallskip

\noindent\textbf{Is \modelab{} still effective if subcategory information is not required?} Tab.~\ref{table:four_class_analysis} we evaluate a setting where the goal is only to predict one of four categories: human written, machine generated, machine paraphrased, and translated (effectively eliminating the subcategories).  We compare a DistilBERT trained to predict these four categories with \modelab{}, where we take the highest subcategory score to represent our confidence in that category.  We see that \modelab{} still obtains 3 mAP gain on average, demonstrating the benefits of leveraging subcategory information even if the fine-grained category predictions are not necessary. 

\begin{figure}[t]
  \includegraphics[width=\columnwidth]{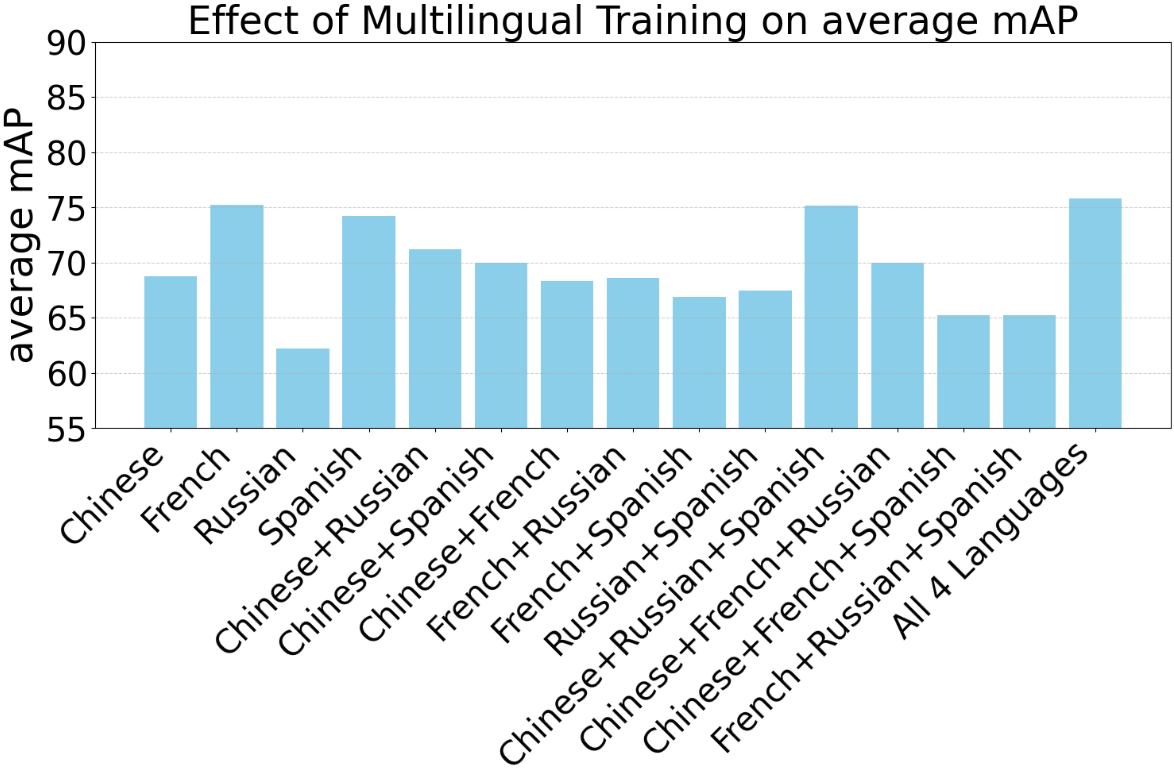}
  \caption{Effect of multilingual training on average mAP across different language combinations evaluated on the four class VisualNews~\citep{liu2020visualnews} setting also reported in Tab.~\ref{table:four_class_analysis}. Models trained on multiple languages generally outperform those trained on a single language, with the highest average mAP observed when training on all four languages.}
  \label{fig:lang_ablation}
\end{figure}

Fig.~\ref{fig:lang_ablation} shows the effect of training on different combinations of languages.  As we increase the number of languages beyond two, we start to see some saturation, where there are smaller differences between models, suggesting that a very large number of languages may not be necessary to recognize a document as originating from another language.

\section{Conclusion}
\label{sec:conclusion}

In this paper, we conduct an in-depth study of fine-grained MGT detection, aiming to further distinguish between machine translated and machine paraphrased texts from MGT. We introduced \modelab{}, a fine-grained machine-influenced text detection framework that goes beyond the classical binary classification approach. Our hierarchical structure, combined with length-specialist models, enables strong generalization across diverse LLMs and varying input lengths, making it suitable for real-world applications. Our extensive experiments across multiple LLMs and different datasets show that \modelab{} consistently outperforms the state-of-the-art by 2.5-3 mAP, and does especially well in out of domain settings. We also show that identifying source languages can boost a model's ability to identify translated text. Overall, \modelab{} enables more accurate detection of machine-influenced content, which is essential for future works in discerning between benign and malicious uses of LLMs.

\noindent\textbf{Acknowledgments.} This material is based upon work supported by DARPA under agreement number HR00112020054. Any opinions, findings, and conclusions or recommendations expressed in this material are those of the author(s) and do not necessarily reflect the views of the supporting agencies.

\section{Limitations}
\label{sec:limitation}
In this paper, we have investigated the FG-MGT task and our proposed \modelab{} shows improved performance over existing detectors. Despite the improved performance, our method still has several limitations, discussed further below.

While our proposed method improves performance for zero-shot evaluations in our experiments, our approach does not guarantee 100\% accuracy on other LLMs and datasets. Therefore, we strongly discourage the use of our approach without proper human supervision (\eg, for plagiarism detection or similar formal applications). A more appropriate application of \modelab{} is to introduce human-supervision for more reliable detection against LLM-generated misinformation.

We also notice the performance difference between in-domain LLM and out-of-domain LLMs. As shown in Sec.~\ref{sec:experiments_finegrainedMGT}, the performance of \modelab{} on out-of-domain generators (StableLM-2, ChatGLM-3, Qwen-2.5, Qwen-1.5) is still lower than that on in-domain generators (Llama-3). Therefore, out-of-domain evaluations remain a challenge for future research in this topic.

Our translated data also utilized a round-trip strategy (discussed in App.~\ref{sec:appendix_round-trip_translation}) to control for content consistency.  However, these translations may also introduce some noise into the articles that may make them easier to detect.  Thus, our results should be seen as only an approximation of the model's true performance.

\bibliography{custom}

\begin{thebibliography}{49}
\providecommand{\natexlab}[1]{#1}

\bibitem[{Abassy et~al.(2024)Abassy, Elozeiri, Aziz, Ta, Tomar, Adhikari, Ahmed, Wang, Afzal, Xie et~al.}]{abassy2024llm}
Mervat Abassy, Kareem Elozeiri, Alexander Aziz, Minh~Ngoc Ta, Raj~Vardhan Tomar, Bimarsha Adhikari, Saad El~Dine Ahmed, Yuxia Wang, Osama~Mohammed Afzal, Zhuohan Xie, and 1 others. 2024.
\newblock Llm-detectaive: a tool for fine-grained machine-generated text detection.
\newblock \emph{arXiv preprint arXiv:2408.04284}.

\bibitem[{Amit et~al.(2004)Amit, Geman, and Fan}]{amit2004coarse}
Yali Amit, Donald Geman, and Xiaodong Fan. 2004.
\newblock A coarse-to-fine strategy for multiclass shape detection.
\newblock \emph{IEEE Transactions on Pattern Analysis and Machine Intelligence}, 26(12):1606--1621.

\bibitem[{Bai et~al.(2023)Bai, Bai, Chu, Cui, Dang, Deng, Fan, Ge, Han, Huang et~al.}]{bai2023qwen}
Jinze Bai, Shuai Bai, Yunfei Chu, Zeyu Cui, Kai Dang, Xiaodong Deng, Yang Fan, Wenbin Ge, Yu~Han, Fei Huang, and 1 others. 2023.
\newblock Qwen technical report.
\newblock \emph{arXiv preprint arXiv:2309.16609}.

\bibitem[{Bao et~al.(2024)Bao, Zhao, Teng, Yang, and Zhang}]{bao2024fast}
Guangsheng Bao, Yanbin Zhao, Zhiyang Teng, Linyi Yang, and Yue Zhang. 2024.
\newblock Fast-detectgpt: Efficient zero-shot detection of machine-generated text via conditional probability curvature.
\newblock In \emph{International Conference on Learning Representations (ICLR)}.

\bibitem[{Bellagente et~al.(2024)Bellagente, Tow, Mahan, Phung, Zhuravinskyi, Adithyan, Baicoianu, Brooks, Cooper, Datta et~al.}]{bellagente2024stable}
Marco Bellagente, Jonathan Tow, Dakota Mahan, Duy Phung, Maksym Zhuravinskyi, Reshinth Adithyan, James Baicoianu, Ben Brooks, Nathan Cooper, Ashish Datta, and 1 others. 2024.
\newblock Stable lm 2 1.6 b technical report.
\newblock \emph{arXiv preprint arXiv:2402.17834}.

\bibitem[{Bhattacharjee et~al.(2023)Bhattacharjee, Kumarage, Moraffah, and Liu}]{bhattacharjee2023conda}
Amrita Bhattacharjee, Tharindu Kumarage, Raha Moraffah, and Huan Liu. 2023.
\newblock Conda: Contrastive domain adaptation for ai-generated text detection.
\newblock In \emph{Proceedings of the 13th International Joint Conference on Natural Language Processing and the 3rd Conference of the Asia-Pacific Chapter of the Association for Computational Linguistics (Volume 1: Long Papers)}, pages 598--610.

\bibitem[{Biten et~al.(2019)Biten, Gomez, Rusinol, and Karatzas}]{biten2019good}
Ali~Furkan Biten, Lluis Gomez, Mar{\c{c}}al Rusinol, and Dimosthenis Karatzas. 2019.
\newblock Good news, everyone! context driven entity-aware captioning for news images.
\newblock In \emph{Proceedings of the IEEE/CVF Conference on Computer Vision and Pattern Recognition}, pages 12466--12475.

\bibitem[{Cao et~al.(2022)Cao, Dong, and Cheung}]{cao2022hallucinated}
Meng Cao, Yue Dong, and Jackie Chi~Kit Cheung. 2022.
\newblock Hallucinated but factual! inspecting the factuality of hallucinations in abstractive summarization.
\newblock In \emph{Proceedings of the 60th Annual Meeting of the Association for Computational Linguistics}, pages 3340--3354.

\bibitem[{Devlin et~al.(2019)Devlin, Chang, Lee, and Toutanova}]{bert}
Jacob Devlin, Ming-Wei Chang, Kenton Lee, and Kristina Toutanova. 2019.
\newblock \href {https://arxiv.org/abs/1810.04805} {Bert: Pre-training of deep bidirectional transformers for language understanding}.
\newblock \emph{Preprint}, arXiv:1810.04805.

\bibitem[{Du et~al.(2022)Du, Qian, Liu, Ding, Qiu, Yang, and Tang}]{du2022glm}
Zhengxiao Du, Yujie Qian, Xiao Liu, Ming Ding, Jiezhong Qiu, Zhilin Yang, and Jie Tang. 2022.
\newblock Glm: General language model pretraining with autoregressive blank infilling.
\newblock In \emph{Proceedings of the 60th Annual Meeting of the Association for Computational Linguistics}.

\bibitem[{Gehrmann et~al.(2019)Gehrmann, Strobelt, and Rush}]{gehrmann2019gltr}
Sebastian Gehrmann, Hendrik Strobelt, and Alexander~M Rush. 2019.
\newblock Gltr: Statistical detection and visualization of generated text.
\newblock In \emph{Proceedings of the 57th Annual Meeting of the Association for Computational Linguistics: System Demonstrations}.

\bibitem[{Gou et~al.(2023)Gou, Shao, Gong, Shen, Yang, Duan, and Chen}]{gou2023critic}
Zhibin Gou, Zhihong Shao, Yeyun Gong, Yelong Shen, Yujiu Yang, Nan Duan, and Weizhu Chen. 2023.
\newblock Critic: Large language models can self-correct with tool-interactive critiquing.
\newblock \emph{arXiv preprint arXiv:2305.11738}.

\bibitem[{Guo et~al.(2023)Guo, Zhang, Wang, Jiang, Nie, Ding, Yue, and Wu}]{guo2023close}
Biyang Guo, Xin Zhang, Ziyuan Wang, Minqi Jiang, Jinran Nie, Yuxuan Ding, Jianwei Yue, and Yupeng Wu. 2023.
\newblock How close is chatgpt to human experts? comparison corpus, evaluation, and detection.
\newblock \emph{arXiv preprint arXiv:2301.07597}.

\bibitem[{Hans et~al.(2024)Hans, Schwarzschild, Cherepanova, Kazemi, Saha, Goldblum, Geiping, and Goldstein}]{hans2024spotting}
Abhimanyu Hans, Avi Schwarzschild, Valeriia Cherepanova, Hamid Kazemi, Aniruddha Saha, Micah Goldblum, Jonas Geiping, and Tom Goldstein. 2024.
\newblock Spotting llms with binoculars: Zero-shot detection of machine-generated text.
\newblock In \emph{International Conference on Machine Learning (ICML)}.

\bibitem[{He et~al.(2021)He, Liu, Gao, and Chen}]{he2021deberta}
Pengcheng He, Xiaodong Liu, Jianfeng Gao, and Weizhu Chen. 2021.
\newblock Deberta: Decoding-enhanced bert with disentangled attention.
\newblock In \emph{International Conference on Learning Representations (ICLR)}.

\bibitem[{He et~al.(2023)He, Shen, Chen, Backes, and Zhang}]{he2023mgtbench}
Xinlei He, Xinyue Shen, Zeyuan Chen, Michael Backes, and Yang Zhang. 2023.
\newblock Mgtbench: Benchmarking machine-generated text detection.
\newblock \emph{arXiv preprint arXiv:2303.14822}.

\bibitem[{Hu et~al.(2023)Hu, Chen, and Ho}]{hu2023radar}
Xiaomeng Hu, Pin-Yu Chen, and Tsung-Yi Ho. 2023.
\newblock \href {https://openreview.net/forum?id=QGrkbaan79} {{RADAR}: Robust {AI}-text detection via adversarial learning}.
\newblock In \emph{Thirty-seventh Conference on Neural Information Processing Systems}.

\bibitem[{Krishna et~al.(2024)Krishna, Song, Karpinska, Wieting, and Iyyer}]{krishna2024paraphrasing}
Kalpesh Krishna, Yixiao Song, Marzena Karpinska, John Wieting, and Mohit Iyyer. 2024.
\newblock Paraphrasing evades detectors of ai-generated text, but retrieval is an effective defense.
\newblock \emph{Advances in Neural Information Processing Systems (NeurIPS)}.

\bibitem[{Kuznetsov et~al.(2024)Kuznetsov, Tulchinskii, Kushnareva, Magai, Barannikov, Nikolenko, and Piontkovskaya}]{kuznetsov-etal-2024-robust}
Kristian Kuznetsov, Eduard Tulchinskii, Laida Kushnareva, German Magai, Serguei Barannikov, Sergey Nikolenko, and Irina Piontkovskaya. 2024.
\newblock Robust {AI}-generated text detection by restricted embeddings.
\newblock In \emph{Findings of the Association for Computational Linguistics: EMNLP 2024}.

\bibitem[{Li et~al.(2024)Li, Wang, Cui, Bi, Shi, and Zhang}]{li2024spotting}
Yafu Li, Zhilin Wang, Leyang Cui, Wei Bi, Shuming Shi, and Yue Zhang. 2024.
\newblock Spotting ai's touch: Identifying llm-paraphrased spans in text.
\newblock In \emph{Findings of the Annual Meeting of the Association for Computational Linguistics: ACL}.

\bibitem[{Lin et~al.(2022)Lin, Hilton, and Evans}]{lin2022truthfulqa}
Stephanie Lin, Jacob Hilton, and Owain Evans. 2022.
\newblock Truthfulqa: Measuring how models mimic human falsehoods.
\newblock In \emph{Proceedings of the 60th Annual Meeting of the Association for Computational Linguistics}, pages 3214--3252.

\bibitem[{Liu et~al.(2021)Liu, Wang, Wang, and Ordonez}]{liu2020visualnews}
Fuxiao Liu, Yinghan Wang, Tianlu Wang, and Vicente Ordonez. 2021.
\newblock Visualnews: Benchmark and challenges in entity-aware image captioning.
\newblock In \emph{Proceedings of the 2021 Conference on Empirical Methods in Natural Language Processing}, pages 6761--6771.

\bibitem[{Liu et~al.(2019)Liu, Ott, Goyal, Du, Joshi, Chen, Levy, Lewis, Zettlemoyer, and Stoyanov}]{liu2019roberta}
Yinhan Liu, Myle Ott, Naman Goyal, Jingfei Du, Mandar Joshi, Danqi Chen, Omer Levy, Mike Lewis, Luke Zettlemoyer, and Veselin Stoyanov. 2019.
\newblock Roberta: A robustly optimized bert pretraining approach.
\newblock \emph{arXiv preprint arXiv:1907.11692}.

\bibitem[{Macko et~al.(2023)Macko, Moro, Uchendu, Lucas, Yamashita, Pikuliak, Srba, Le, Lee, Simko, and Bielikova}]{macko-etal-2023-multitude}
Dominik Macko, Robert Moro, Adaku Uchendu, Jason Lucas, Michiharu Yamashita, Mat{\'u}{\v{s}} Pikuliak, Ivan Srba, Thai Le, Dongwon Lee, Jakub Simko, and Maria Bielikova. 2023.
\newblock {MULTIT}u{DE}: Large-scale multilingual machine-generated text detection benchmark.
\newblock In \emph{Proceedings of the 2023 Conference on Empirical Methods in Natural Language Processing}.

\bibitem[{Mao et~al.(2025)Mao, Zhang, Zhang, and Zhao}]{10888686}
Dianhui Mao, Denghui Zhang, Ao~Zhang, and Zhihua Zhao. 2025.
\newblock Mlsdet: Multi-llm statistical deep ensemble for chinese ai-generated text detection.
\newblock In \emph{ICASSP 2025 - 2025 IEEE International Conference on Acoustics, Speech and Signal Processing (ICASSP)}.

\bibitem[{Maynez et~al.(2020)Maynez, Narayan, Bohnet, and McDonald}]{maynez2020faithfulness}
Joshua Maynez, Shashi Narayan, Bernd Bohnet, and Ryan McDonald. 2020.
\newblock On faithfulness and factuality in abstractive summarization.
\newblock In \emph{Proceedings of the 58th Annual Meeting of the Association for Computational Linguistics}, pages 1906--1919.

\bibitem[{Meng et~al.(2022)Meng, Bau, Andonian, and Belinkov}]{meng2022locating}
Kevin Meng, David Bau, Alex Andonian, and Yonatan Belinkov. 2022.
\newblock Locating and editing factual associations in gpt.
\newblock \emph{Advances in Neural Information Processing Systems}, 35:17359--17372.

\bibitem[{Miralles-González et~al.(2025)Miralles-González, Huertas-Tato, Martín, and Camacho}]{mirallesgonzález2025tokenscreatedequalperplexity}
Pablo Miralles-González, Javier Huertas-Tato, Alejandro Martín, and David Camacho. 2025.
\newblock \href {https://arxiv.org/abs/2501.03940} {Not all tokens are created equal: Perplexity attention weighted networks for ai generated text detection}.
\newblock \emph{Preprint}, arXiv:2501.03940.

\bibitem[{Mitchell et~al.(2023)Mitchell, Lee, Khazatsky, Manning, and Finn}]{mitchell2023detectgpt}
Eric Mitchell, Yoonho Lee, Alexander Khazatsky, Christopher~D Manning, and Chelsea Finn. 2023.
\newblock Detectgpt: Zero-shot machine-generated text detection using probability curvature.
\newblock In \emph{International Conference on Machine Learning (ICML)}.

\bibitem[{Nguyen-Son et~al.(2021)Nguyen-Son, Thao, Hidano, Gupta, and Kiyomoto}]{nguyen2021machine}
Hoang-Quoc Nguyen-Son, Tran Thao, Seira Hidano, Ishita Gupta, and Shinsaku Kiyomoto. 2021.
\newblock Machine translated text detection through text similarity with round-trip translation.
\newblock In \emph{Proceedings of the 2021 Conference of the North American Chapter of the Association for Computational Linguistics: Human Language Technologies}, pages 5792--5797.

\bibitem[{Ouyang et~al.(2022)Ouyang, Wu, Jiang, Almeida, Wainwright, Mishkin, Zhang, Agarwal, Slama, Ray et~al.}]{ouyang2022training}
Long Ouyang, Jeffrey Wu, Xu~Jiang, Diogo Almeida, Carroll Wainwright, Pamela Mishkin, Chong Zhang, Sandhini Agarwal, Katarina Slama, Alex Ray, and 1 others. 2022.
\newblock Training language models to follow instructions with human feedback.
\newblock \emph{Advances in Neural Information Processing Systems (NeurIPS)}.

\bibitem[{Parikh et~al.(2020)Parikh, Wang, Gehrmann, Faruqui, Dhingra, Yang, and Das}]{parikh2020totto}
Ankur Parikh, Xuezhi Wang, Sebastian Gehrmann, Manaal Faruqui, Bhuwan Dhingra, Diyi Yang, and Dipanjan Das. 2020.
\newblock Totto: A controlled table-to-text generation dataset.
\newblock In \emph{Proceedings of the 2020 Conference on Empirical Methods in Natural Language Processing (EMNLP)}, pages 1173--1186.

\bibitem[{Radford et~al.(2019)Radford, Wu, Child, Luan, Amodei, Sutskever et~al.}]{radford2019language}
Alec Radford, Jeffrey Wu, Rewon Child, David Luan, Dario Amodei, Ilya Sutskever, and 1 others. 2019.
\newblock Language models are unsupervised multitask learners.
\newblock \emph{OpenAI blog}, 1(8):9.

\bibitem[{Sanh et~al.(2020)Sanh, Debut, Chaumond, and Wolf}]{distilbert}
Victor Sanh, Lysandre Debut, Julien Chaumond, and Thomas Wolf. 2020.
\newblock \href {https://arxiv.org/abs/1910.01108} {Distilbert, a distilled version of bert: smaller, faster, cheaper and lighter}.
\newblock \emph{Preprint}, arXiv:1910.01108.

\bibitem[{Shuster et~al.(2021)Shuster, Poff, Chen, Kiela, and Weston}]{shuster2021retrieval}
Kurt Shuster, Spencer Poff, Moya Chen, Douwe Kiela, and Jason Weston. 2021.
\newblock Retrieval augmentation reduces hallucination in conversation.
\newblock In \emph{Findings of the Association for Computational Linguistics: EMNLP 2021}, pages 3784--3803.

\bibitem[{Solaiman et~al.(2019)Solaiman, Brundage, Clark, Askell, Herbert-Voss, Wu, Radford, Krueger, Kim, Kreps et~al.}]{solaiman2019release}
Irene Solaiman, Miles Brundage, Jack Clark, Amanda Askell, Ariel Herbert-Voss, Jeff Wu, Alec Radford, Gretchen Krueger, Jong~Wook Kim, Sarah Kreps, and 1 others. 2019.
\newblock Release strategies and the social impacts of language models.
\newblock \emph{arXiv preprint arXiv:1908.09203}.

\bibitem[{Stephen et~al.(2017)Stephen, Caiming, James, and Socher}]{merity2016pointer}
Merity Stephen, Xiong Caiming, Bradbury James, and Richard Socher. 2017.
\newblock Pointer sentinel mixture models.
\newblock \emph{Proceedings of ICLR}.

\bibitem[{Su et~al.(2023)Su, Zhuo, Wang, and Nakov}]{su2023detectllm}
Jinyan Su, Terry~Yue Zhuo, Di~Wang, and Preslav Nakov. 2023.
\newblock \href {https://openreview.net/forum?id=Dy2mbQIdMz} {Detect{LLM}: Leveraging log rank information for zero-shot detection of machine-generated text}.
\newblock In \emph{Findings of the Association for Computational Linguistics: EMNLP}.

\bibitem[{Tian and Cui(2023)}]{tian2023gptzero}
Edward Tian and Alexander Cui. 2023.
\newblock \href {https://gptzero.me} {Gptzero: Towards detection of ai-generated text using zero-shot and supervised methods}.

\bibitem[{Tian et~al.(2024)Tian, Chen, Wang, Bai, Zhang, Li, Xu, and Wang}]{tian2024multiscale}
Yuchuan Tian, Hanting Chen, Xutao Wang, Zheyuan Bai, Qinghua Zhang, Ruifeng Li, Chao Xu, and Yunhe Wang. 2024.
\newblock \href {https://openreview.net/forum?id=5Lp6qU9hzV} {Multiscale positive-unlabeled detection of {AI}-generated texts}.
\newblock In \emph{The Twelfth International Conference on Learning Representations}.

\bibitem[{Touvron et~al.(2023)Touvron, Lavril, Izacard, Martinet, Lachaux, Lacroix, Rozi{\`e}re, Goyal, Hambro, Azhar, Rodriguez, Joulin, Grave, and Lample}]{touvron2023llama}
Hugo Touvron, Thibaut Lavril, Gautier Izacard, Xavier Martinet, Marie-Anne Lachaux, Timoth{\'e}e Lacroix, Baptiste Rozi{\`e}re, Naman Goyal, Eric Hambro, Faisal Azhar, Aurelien Rodriguez, Armand Joulin, Edouard Grave, and Guillaume Lample. 2023.
\newblock Llama: Open and efficient foundation language models.
\newblock \emph{arXiv preprint arXiv:2302.13971}.

\bibitem[{Verma et~al.(2024)Verma, Fleisig, Tomlin, and Klein}]{verma2024ghostbuster}
Vivek Verma, Eve Fleisig, Nicholas Tomlin, and Dan Klein. 2024.
\newblock Ghostbuster: Detecting text ghostwritten by large language models.
\newblock In \emph{North American Chapter of the Association for Computational Linguistics (NAACL)}.

\bibitem[{Wang et~al.(2020)Wang, Qinami, Karakozis, Genova, Nair, Hata, and Russakovsky}]{Wang_2020_CVPR}
Zeyu Wang, Klint Qinami, Ioannis Karakozis, Kyle Genova, Prem Nair, Kenji Hata, and Olga Russakovsky. 2020.
\newblock Towards fairness in visual recognition: Effective strategies for bias mitigation.
\newblock In \emph{Proceedings of the IEEE/CVF Conference on Computer Vision and Pattern Recognition (CVPR)}.

\bibitem[{Xu et~al.(2023)Xu, Ding, Wang, Yang, Yu, Yu, and Xia}]{xu2023dynamic}
Chang Xu, Jian Ding, Jinwang Wang, Wen Yang, Huai Yu, Lei Yu, and Gui-Song Xia. 2023.
\newblock Dynamic coarse-to-fine learning for oriented tiny object detection.
\newblock In \emph{Proceedings of the IEEE/CVF Conference on Computer Vision and Pattern Recognition}, pages 7318--7328.

\bibitem[{Yang et~al.(2024)Yang, Yang, Hui, Zheng, Yu, Zhou, Li, Li, Liu, Huang et~al.}]{yang2024qwen2}
An~Yang, Baosong Yang, Binyuan Hui, Bo~Zheng, Bowen Yu, Chang Zhou, Chengpeng Li, Chengyuan Li, Dayiheng Liu, Fei Huang, and 1 others. 2024.
\newblock Qwen2 technical report.
\newblock \emph{arXiv preprint arXiv:2407.10671}.

\bibitem[{Yuan et~al.(2023)Yuan, Cheng, Yan, Zeng, and Han}]{yuan2023small}
Xiang Yuan, Gong Cheng, Kebing Yan, Qinghua Zeng, and Junwei Han. 2023.
\newblock Small object detection via coarse-to-fine proposal generation and imitation learning.
\newblock In \emph{Proceedings of the IEEE/CVF international conference on computer vision}, pages 6317--6327.

\bibitem[{Zellers et~al.(2019)Zellers, Holtzman, Rashkin, Bisk, Farhadi, Roesner, and Choi}]{zellers2019defending}
Rowan Zellers, Ari Holtzman, Hannah Rashkin, Yonatan Bisk, Ali Farhadi, Franziska Roesner, and Yejin Choi. 2019.
\newblock Defending against neural fake news.
\newblock \emph{Advances in neural information processing systems (NeurIPS)}, 32.

\bibitem[{Zhang et~al.(2024)Zhang, Qin, and Plummer}]{zhang2024MGTL}
Zhongping Zhang, Wenda Qin, and Bryan~A. Plummer. 2024.
\newblock Machine-generated text localization.
\newblock In \emph{Findings of the Annual Meeting of the Association for Computational Linguistics: ACL}.

\bibitem[{Zhou et~al.(2021)Zhou, Neubig, Gu, Diab, Guzm{\'a}n, Zettlemoyer, and Ghazvininejad}]{zhou2021detecting}
Chunting Zhou, Graham Neubig, Jiatao Gu, Mona Diab, Francisco Guzm{\'a}n, Luke Zettlemoyer, and Marjan Ghazvininejad. 2021.
\newblock Detecting hallucinated content in conditional neural sequence generation.
\newblock In \emph{Findings of the Association for Computational Linguistics: ACL-IJCNLP 2021}, pages 1393--1404.

\end{thebibliography}

\newpage
\appendix

\section*{Appendix}
\label{sec:appendix}
\section{Additional Results}
\label{sec:additional_results}

\begin{figure}[t]
  \includegraphics[width=\columnwidth]{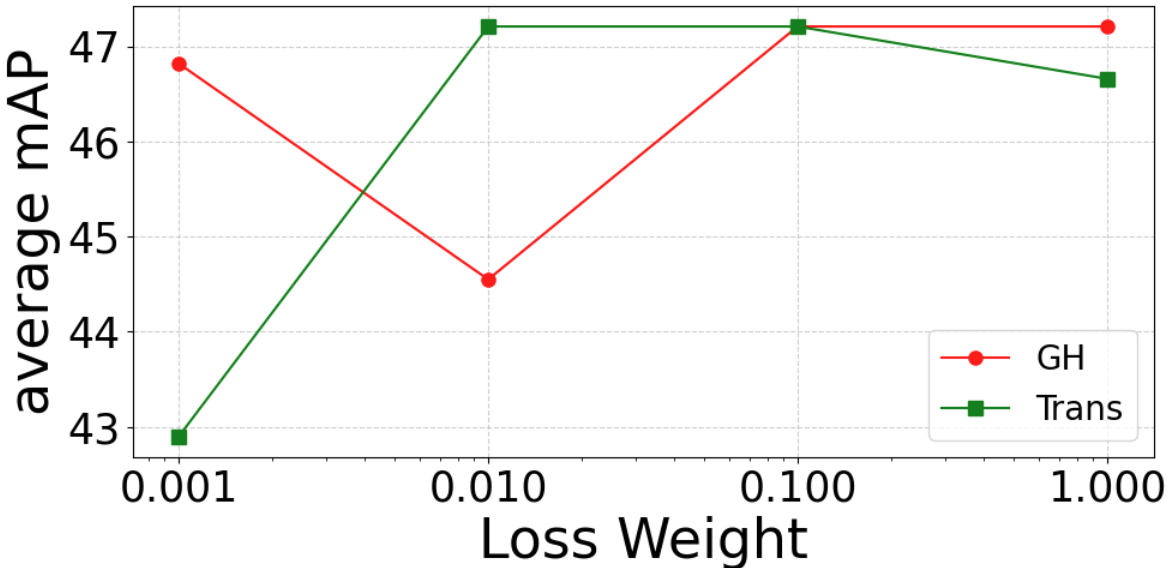}
  \caption{Effect of GH (Generate-Humanized) and Trans loss weights for guided learning on average mAP performance evaluated on VisualNews~\citep{liu2020visualnews}. The model achieves the highest mAP when both the GH and Trans loss weights are set to 0.01.}
  \label{fig:guided_learning}
\end{figure}

Fig.~\ref{fig:guided_learning} shows the effect of changing the loss weight $\lambda$ from Eq.~\ref{eq:guided}.  The same value of $\lambda$ performs best for both, reducing the number of hyperparameters that need to be tuned for our model.

\begin{figure}[t]
  \includegraphics[width=\columnwidth]{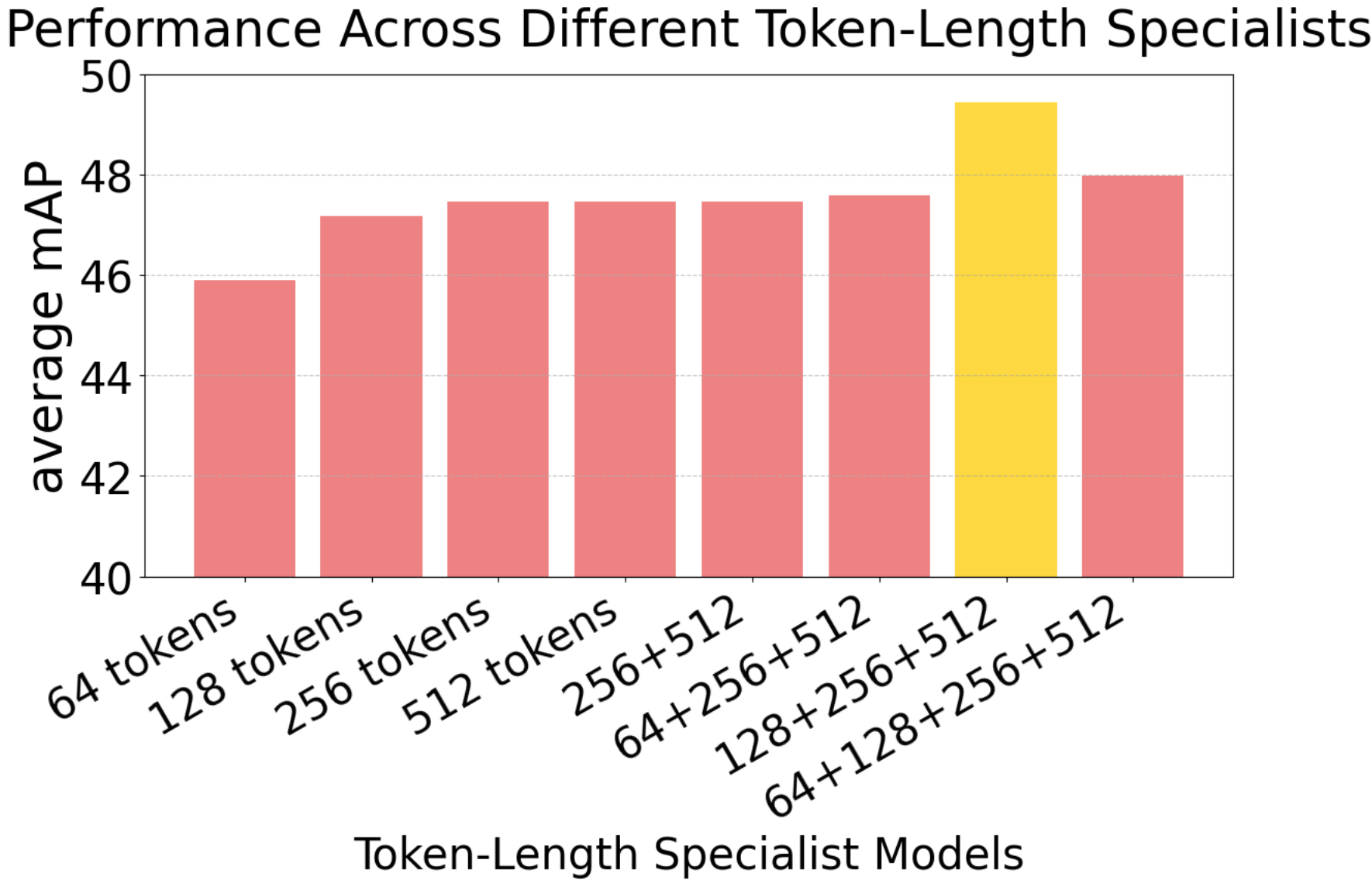}
  \caption{Average mAP across different token-length specialist models evaluated on VisualNews~\citep{liu2020visualnews}. Models trained with a single token length achieve moderate performance, while combining specialists across multiple token lengths significantly improves detection accuracy. The highest average mAP is observed when using specialists for 128, 256, and 512 tokens.}
  \label{fig:ensemble}
\end{figure}

Fig.~\ref{fig:ensemble} ablates the number and size of experts to train. We find that three experts generally provide enough coverage to perform well on a diverse set of lengths.  That said, the number of experts likely would vary depending on the maximum input sequence a model can support.  However, very long documents are easier to detect as machine-generated (see Tab.~\ref{table:token_lengths_analysis}), so support for very long sequences may not be necessary as a model may be able to effectively detect a language model was used on just part of a document.

\begin{figure}[t]
    \centering
    \includegraphics[width=\linewidth]{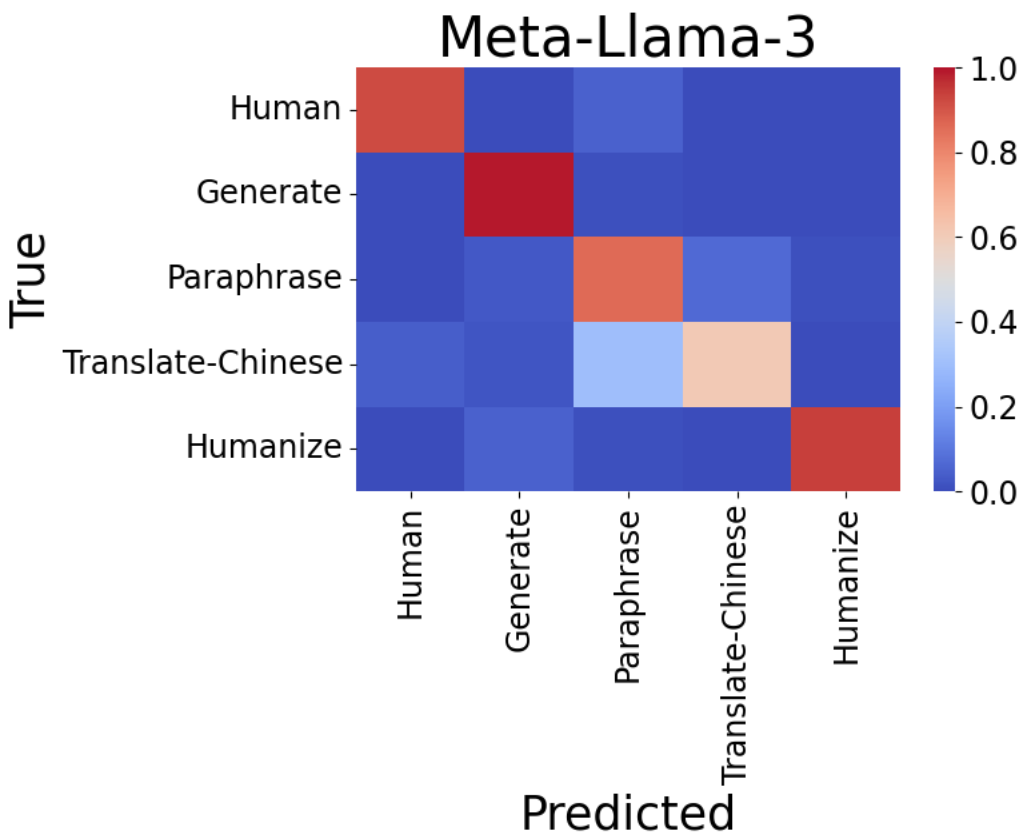}
    \vspace{-4mm}
    \caption{Confusion Matrix for in-domain LLMs on VisualNews. \modelab{} performs well in most categories, especially on the machine-translated articles.}
    \vspace{-2mm}
    \label{fig:confusion_matrix_indomain}
\end{figure}

\begin{figure*}[t]
    \centering
    \includegraphics[width=\linewidth]{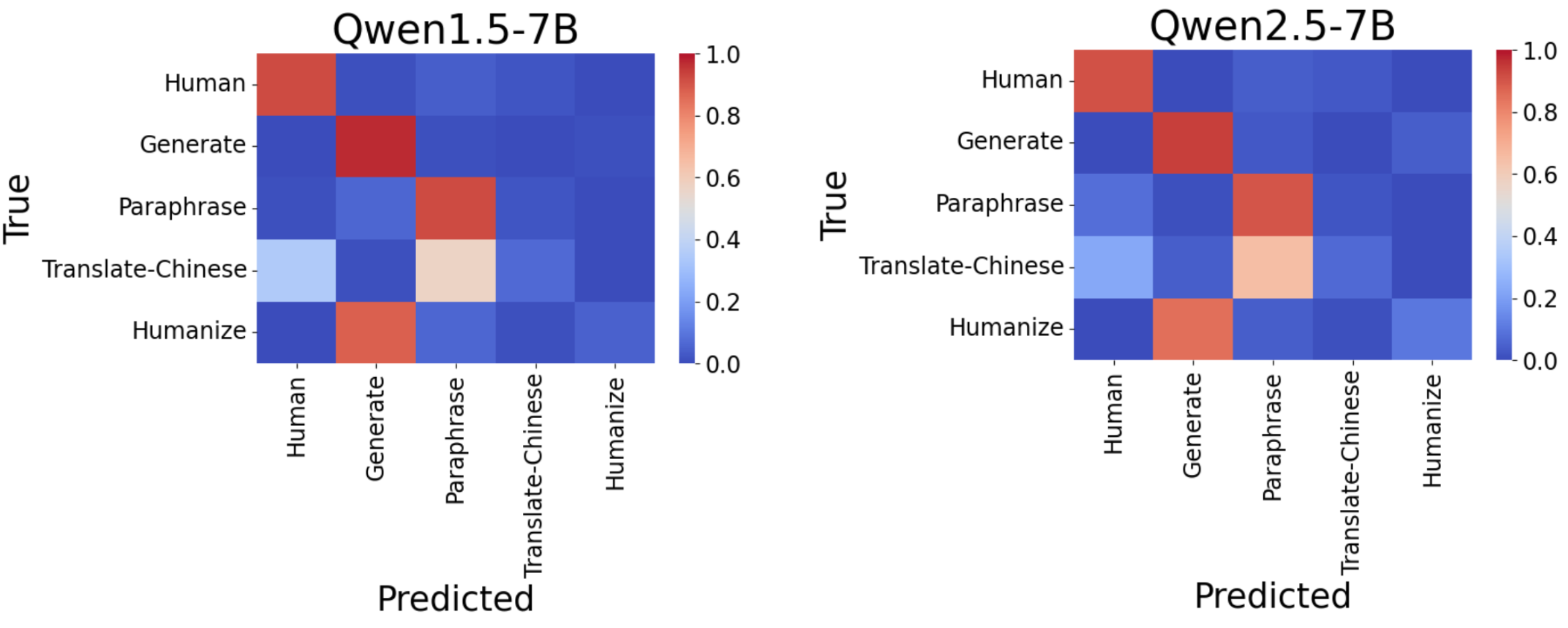}
    \caption{Confusion Matrix on out-of-domain LLMs on VisualNews. Our method can still accurately distinguish between human-written and machine-generated categories. However, when compared to in-domain evaluations in Fig.~\ref{fig:confusion_matrix_indomain}, detecting machine-humanized text becomes more challenging.}
    \label{fig:confusion_matrix_outofdomain}
\end{figure*}

\noindent \textbf{Class confusion matrix.} To provide a more intuitive understanding of  \modelab{}, we provide the visualization for \modelab{}'s performance across different FG-MGT categories on VisualNews using confusion matrices as shown in Fig.~\ref{fig:confusion_matrix_indomain} and Fig.~\ref{fig:confusion_matrix_outofdomain}. The results show that \modelab{} can accurately distinguish translated text from different source languages, even when evaluated on out-of-domain LLMs. However, the model continues to struggle with distinguishing between generated and humanized content. This challenge may stem from the fact that both types are produced by LLMs using human written input, resulting in similar surface-level characteristics.

\begin{figure*}[t]
    \centering
    \includegraphics[width=0.6\linewidth]{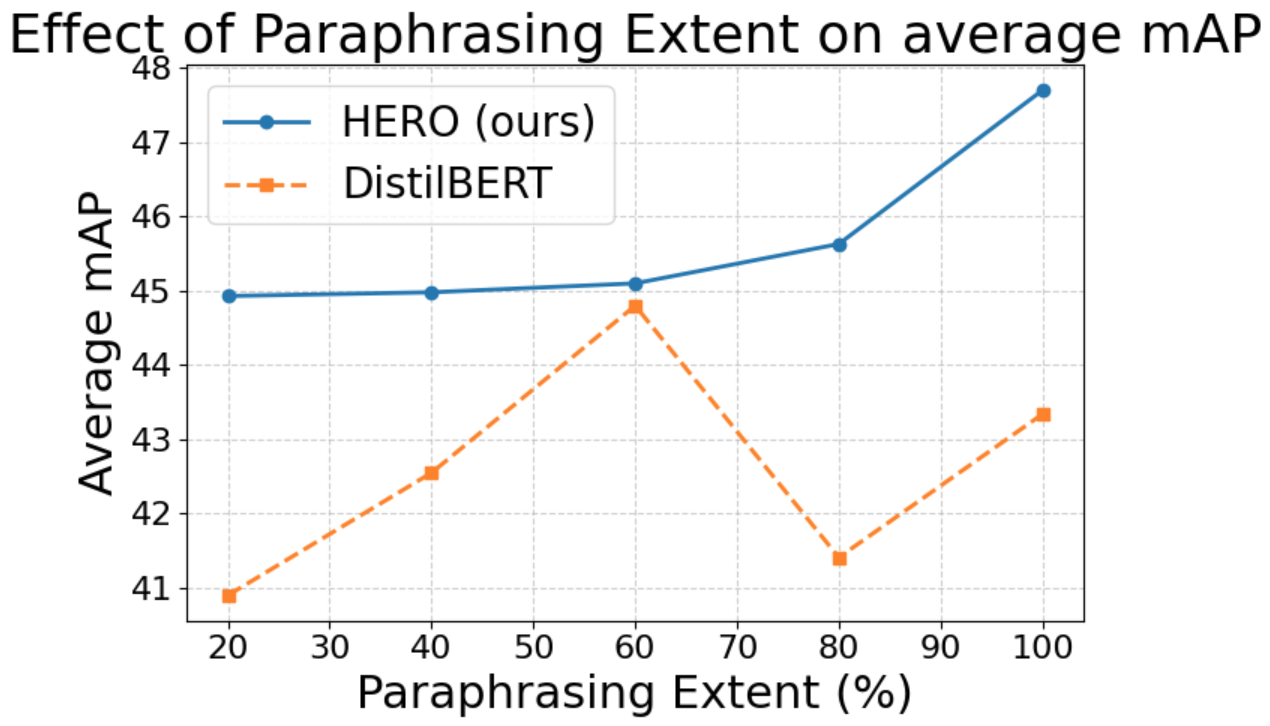}
    \caption{We illustrate the effect of paraphrasing extent on average mAP. Higher levels of paraphrasing improve the model's performance.}
    \label{fig:paraphrase_extent}
\end{figure*}

\section{Effect of Paraphrasing Extent}
\label{sec:appendix_paraphrase}

To examine how the extent of paraphrasing can affect the performance of \modelab, we paraphrased 20\%, 40\%, 60\%, 80\%, and 100\% of the input text. The resulting performance is shown in Fig.~\ref{fig:paraphrase_extent}. As paraphrasing extent increases, the detection accuracy also improves, suggesting that higher levels of paraphrasing make manipulation patterns more discernible to the model.

\section{Round-trip Translation Strategy}
\label{sec:appendix_round-trip_translation}

\begin{figure*}[t]
    \centering
    \includegraphics[width=1.0\linewidth]{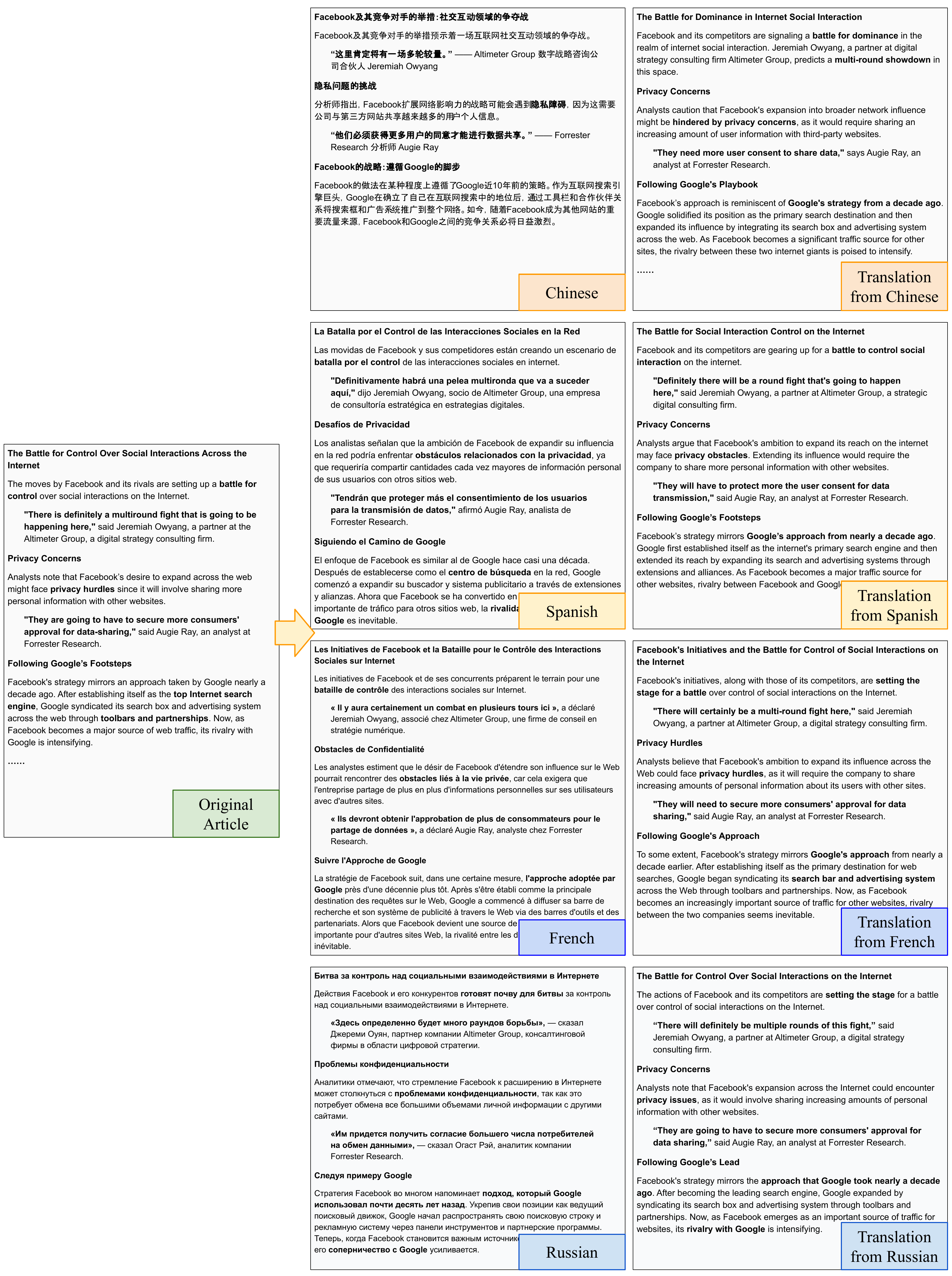}
    \caption{\textbf{Round-trip Strategy for Generating Translated Articles.}This strategy allows us to automatically produce translated articles from existing datasets, eliminating the need for additional data collection. See Sec.~\ref{sec:appendix_round-trip_translation} for discussion.
    }
    \label{fig:appendix_roundtrip_translation}
\end{figure*}

To create translated versions of the same documents, we adopt the strategy of round-trip translation to generate translated data for our FG-MGT task. Fig.~\ref{fig:appendix_roundtrip_translation} provides a specific example: we first translate the original article into target languages (Chinese, Spanish, French, Russian), and then translate these articles back into English, obtaining machine-translated articles for detection.

\section{Humanized example}
\label{sec:appendix_humanized}

\begin{figure*}[t]
    \centering
    \includegraphics[width=0.8\linewidth]{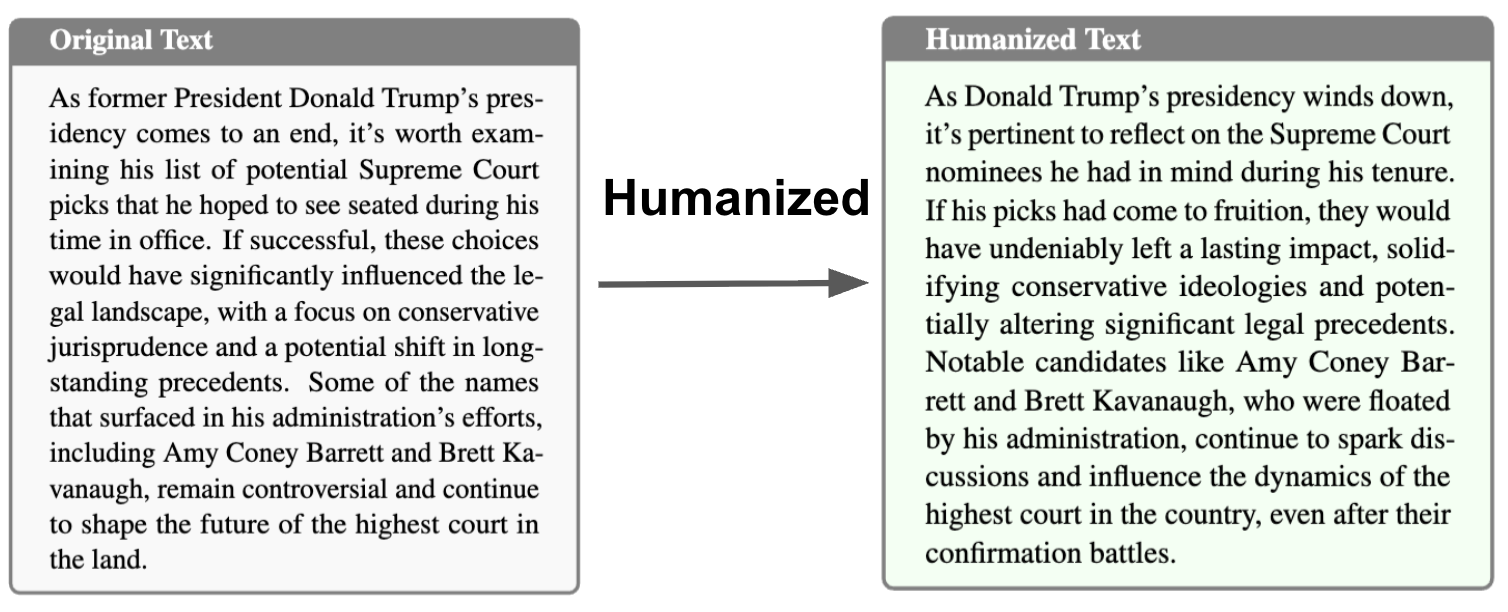}
    \caption{\textbf{Humanized text example}. We utilize machine-generated text and ask the LLMs to rewrite it to sound more natural and human-like, 
while maintaining the same level of detail and length.
    }
    \label{fig:humanized_example}
\end{figure*}

The purpose of machine-humanized data is to simulate a setting where a bad actor may attempt to make their generated text harder to detect.  It accomplishes this by querying a LLM with a request to make the input article sound more human using processes based on those from \citet{abassy2024llm} (see Sec.~\ref{sec:generation} for additional discussion).  Fig.~\ref{fig:humanized_example} shows an example of the differences produced by machine-humanized data.

\end{document}